\documentclass[journal]{IEEEtran}
\usepackage{amsmath,amsfonts}
\usepackage{algorithmic}
\usepackage{algorithm}
\usepackage{array}
\usepackage[caption=false,font=normalsize,labelfont=sf,textfont=sf]{subfig}
\usepackage{textcomp}
\usepackage{stfloats}
\usepackage{url}
\usepackage{verbatim}
\usepackage{graphicx}
\usepackage{cite}
\usepackage{amssymb,mathrsfs,bm}
\usepackage{booktabs}
\usepackage[capitalise]{cleveref}
\usepackage{float}
\usepackage{fp,tikz,pgfplots}
\usepackage{verbatim}
\usepackage{makecell}
\usepackage{multirow}
\usepackage{enumitem}
\usepackage{lipsum} 
\usepackage{chngcntr}

\newtheorem{theorem}{Theorem}
\newtheorem{remark}{Remark}
\newtheorem{lemma}{Lemma}
\newtheorem{property}{Property}
\newtheorem{assumption}{Assumption}

\usepackage{xcolor}
\xdefinecolor{blueZ}{RGB}{6,154,243}
\xdefinecolor{redZ}{RGB}{232,61,117}

\DeclareMathOperator*{\argmin}{arg\,min}

\newcommand{\SWITCH}[1]{\STATE \textbf{switch} (#1)}
\newcommand{\ENDSWITCH}{\STATE \textbf{end switch}}
\newcommand{\CASE}[1]{\STATE \textbf{case} #1\textbf{:} \begin{ALC@g}}
\newcommand{\ENDCASE}{\end{ALC@g}}

\begin{document}
%
\title{Quality or Quantity? Error-Informed Selective Online Learning with Gaussian Processes in Multi-Agent Systems: Extended Version
}

\author{
Zewen Yang,~\IEEEmembership{Member,~IEEE}, Xiaobing Dai, Jiajun Cheng, Yulong Huang,~\IEEEmembership{Senior Member,~IEEE}, Peng Shi,~\IEEEmembership{Fellow,~IEEE}

\thanks{This work was supported by the Federal Ministry of Education and Research of Germany in the programme of ``Souverän. Digital. Vernetzt.'' under joint project 6G-life with project identification number: 16KISK002, and the National Natural Science Foundation of China under Grant Numbers U24B20184 and 62373118. \textit{(Corresponding author: Xiaobing Dai; Yulong Huang.)}}
\thanks{Z. Yang is with the Chair of Robotics and Systems Intelligence (RSI), Munich Institute of Robotics and Machine Intelligence (MIRMI), Technical University of Munich (TUM), 80992 Munich, Germany (e-mail:zewen.yang@tum.de).}
\thanks{X. Dai is with the School of Computation, Information and Technology (CIT), Technical University of Munich (TUM), 80333 Munich, Germany (email: xiaobing.dai@tum.de).}
\thanks{J. Cheng and Y. Huang are with the College of Intelligent Systems Science and Engineering, Harbin Engineering University, Harbin 150001, China (e-mail: chengjiajun@hrbeu.edu.cn; huangyl@hrbeu.edu.cn).}
\thanks{P. Shi is with the School of Electrical and Mechanical Engineering, the University of Adelaide, Adelaide, SA 5005, Australia, and also with the Research and Innovation Centre, Obuda University, 1034 Budapest, Hungary (e-mail: peng.shi@adelaide.edu.au).}
}



\maketitle

\begin{abstract}
Effective cooperation is pivotal in distributed learning for multi-agent systems, where the interplay between the quantity and quality of the machine learning models is crucial. This paper reveals the irrationality of indiscriminate inclusion of all models on agents for joint prediction, highlighting the imperative to prioritize quality over quantity in cooperative learning. 
Specifically, we present the first selective online learning framework for distributed Gaussian process (GP) regression, namely distributed error-informed GP (EIGP), that enables each agent to assess its neighboring collaborators, using the proposed selection function to choose the higher quality GP models with less prediction errors. 
Moreover, algorithmic enhancements are embedded within the EIGP, including a greedy algorithm (gEIGP) for accelerating prediction and an adaptive algorithm (aEIGP) for improving prediction accuracy. In addition, approaches for fast prediction and model update are introduced in conjunction with the error-informed quantification term iteration and a data deletion strategy to achieve real-time learning operations. Numerical simulations are performed to demonstrate the effectiveness of the developed methodology, showcasing its superiority over the state-of-the-art distributed GP methods with different benchmarks.
\end{abstract}


%
\IEEEpeerreviewmaketitle

\section{Introduction}
\IEEEPARstart{T}{he} {investigation of distributed dynamical systems has attracted considerable attention, encompassing a broad range of applications, such as distributed generation systems, traffic networks, networked Internet of Things devices, and multi-robot systems~\cite{yang2024cooperativeF,wang2023event,Friha2021Internet,yan2020virtual}.
However, much of the existing research assumes that system dynamics are fully known, often overlooking uncertainties arising from modeling inaccuracies and environmental disturbances~\cite{li2024dynamic}.
To address these challenges, machine learning techniques offer a promising approach, enabling the discovery of hidden patterns and the inference of unknown system relationships.}

In the context of multi-agent systems (MASs), while each agent can make independent inferences using a local model, this approach lacks fault tolerance, struggles with large datasets, and is prone to biases due to the partial distribution of the dataset. 
Cooperative learning among agents offers a flexible solution to these limitations by sharing useful information, where agents can compensate for each other’s errors, verify prediction consistency, and enhance the prediction accuracy~\cite{ma2024finite,zhang2023privacy,wang2022distributed}.
Among the various techniques employed in cooperative learning, Gaussian Process (GP) regression has gained significant attention due to its non-parametric characteristics, which is well-suited for real-time online inference and monitoring~\cite{wang2021speed,liu2024robust}. 
Additionally, the incorporation of prior knowledge into the data-driven model capturing physical model structures further enhances its interpretability across diverse MASs~\cite{zhu2024multilayer}. 
Moreover, GP regression provides prediction error bounds, offering valuable insights into the reliability of model outputs~\cite{rasmussenGaussianProcessesMachine2006}. 
The predominant distributed Gaussian process methodologies inclusively integrate all local models (experts), overlooking the pivotal consideration of model selection~\cite{liuWhenGaussianProcess2020}. 
The indiscriminate aggregation of all GP models presents an inherent risk, particularly in instances where certain models exhibit inaccuracies, for instance, the mixture of experts (MOE) method~\cite{trespMixturesGaussianProcesses2000}, wherein each model's prediction is accorded equal weight, potentially leading to inadequate results. 
To address this limitation, various techniques have been explored to enhance aggregate weights, such as the product of experts (POE), which leverages the posterior variance of GP models~\cite{hintonTrainingProductsExperts2002}, and the generalized product of experts (GPOE), incorporating the difference in differential entropy between the prior and posterior~\cite{caoGeneralizedProductExperts2015}. 
In contrast to the POE family, the Bayesian committee machine (BCM)~\cite{trespBayesianCommitteeMachine2000} and robust BCM (RBCM)~\cite{deisenrothDistributedGaussianProcesses2015}
incorporates the GP prior, yet exhibits limitations in providing consistent predictions as the training dataset continuously expands.
The generalized RBCM, which necessitates the sharing of datasets among models, however, poses challenges in the context of online learning~\cite{liu2018generalized}.

Recent efforts have explored distributed GP methodologies for MASs. For example, some works leverage the POE to refine aggregation weights~\cite{yangDistributedLearningConsensus2021,yang2024cooperative}, while others employ the dynamical average consensus (DAC) algorithm to reduce the communication burdens~\cite{ledererCooperativeControlUncertain2023,yuan2024lightweight}. 
Nevertheless, these endeavors fall short in consideration of online learning problems. 
To address this limitation, \cite{hoangCollectiveOnlineLearning2019,heDistributedOnlineSparse2023,dai2023can,yang2025streaminggeneratedgaussianprocess} have filled this critical gap, emphasizing the necessity of integrating online data to ensure adaptability in dynamic environments. 
Building on this, \cite{dai2024decentralized,dai2024cooperative} introduce event-triggered mechanisms to improve the efficiency of data acquisition and storage management.
Additionally, to construct environmental maps more effectively, approaches such as distributed sparse GP~\cite{Ding_RAL2024_Resource} and distributed active learning~\cite{jang_RAL2020_multi} have been proposed in an online cooperative learning setting.
However, these works assume that all local models are of equal quality in terms of prediction accuracy, which may not hold in practical cooperative learning scenarios. 
The research~\cite{yang2024whom} introduced ``Pri-GP'', an offline elective distributed GP algorithm designed to evaluate prior model errors. However, Pri-GP's efficacy is contingent upon the variance of agents' prior models, frequently assuming an unknown (zero-mean prior) in common scenarios. Consequently, this methodology proves ineffective when all agents share identical prior models. Furthermore, Pri-GP adopts a static and predetermined selection of agents, lacking adaptive capability.

Therefore, our primary goal is to enable {scalable, reliable, real-time cooperative learning} in MASs with three critical constraints, while addressing three key challenges:
(1) mitigating the computational burden of centralized GP models, which scale as $\mathcal{O}(N^3)$ with dataset size; (2) filtering out unreliable predictions from neighboring agents to ensure accurate joint predictions; (3) enabling continuous model updates with streaming data for online adaptability. 
In response to the challenges, we propose an approach named distributed error-informed Gaussian processes (EIGP), representing, to the best of our knowledge, the first-of-its-kind cooperative online learning method with selective model functionality.
The major contributions of this paper are fourfold. 
Firstly, we introduce a novel error-informed quantifiable metric for evaluating GP models, enabling agents to judiciously select neighboring collaborators. 
Secondly, leveraging the proposed EIGP framework, we devise the greedy EIGP algorithm to empower agents in selectively choosing the optimal collaborator. 
Additionally, an adaptive EIGP algorithm is introduced, allowing each agent to choose neighbors with a specified confidence level. 
Both approaches alleviate computational burdens associated with unnecessary interactions with all neighbors and enhance individual agent predictions by excluding potentially misleading predictions from low-quality models.
Thirdly, we address the challenges of infinite data storage requirements in real-time online learning through a data deletion strategy and continuous model evaluations facilitated by computationally efficient update and prediction methods.
Lastly, we present a rigorous theoretical analysis, establishing a reliable prediction error bound that ensures guaranteed cooperative inference.

The paper is structured as follows: In \cref{sec_ProblemState}, we illustrate the objectives of our study and present foundational concepts, encompassing graph theory and GP regression. Subsequently, in \cref{sec_EIGP}, we introduce the EIGP framework alongside the real-time model update strategy and evaluate its prediction performance. \cref{sec_sim} is dedicated to empirical simulations, wherein we substantiate the superiority of our proposed approach over the existing state-of-the-art distributed GP methods. Finally, conclusions are drawn in  \cref{sec_conclusion}.

\section{Preliminaries and Problem Statement}
\label{sec_ProblemState}
\subsection{Notation and Graph Theory}
\subsubsection{Notation}
We denote $\mathbb{R}$ by the set of real numbers and let $\mathbb{R}_+$ represents positive reals $(0,\infty)$ and $\mathbb{R}_{0,+}$ non-negative reals $[0,\infty)$. 
Natural numbers are symbolized by $\mathbb{N}$ and $\mathbb{N}_+$ without zero.
The $N\times N$ identity matrix as $\bm{I}_N$. 
Define the function $\underline{\lambda}(\cdot)$ and $\bar{\lambda}(\cdot)$ return the minimal and maximal eigenvalue of the matrix.
Additionally, we represent a diagonal matrix as $\mathrm{diag}(a_1, \dots, a_N)$. 
$|\bullet|$ denotes the cardinality of the set $\bullet$.

\subsubsection{Graph Theory} To describe the information exchange of MASs with $n \in \mathbb{N}$ agents, an undirected graph $\mathcal{G} = (\mathcal{V}, \mathcal{E})$ is utilized. The representation of the agents' indices is denoted by the node set $\mathcal{V} = \{1, \ldots, n\}$, while $\mathcal{E} \subseteq \mathcal{V} \times \mathcal{V}$ denotes the collection of edges connecting these nodes. For an edge $\left ( i, j \right ) \in \mathcal{E}$, it indicates information exchange between agent $i$ and agent $j$, and vice versa. We let the set of neighbors of agent $i$ denote $\mathcal{N}_i= \{j\in \mathcal{V}:(i,j)\in {\mathcal{E}}\}$ encompassing its neighboring agents and define its self-included set $\bar{\mathcal{N}}_i = \{i, \mathcal{N}_i\}$.

\subsection{Objective}
In this paper, we aim to estimate a multi-dimensional unknown function $\boldsymbol{f}(\boldsymbol{x}) = [{f}^1(\boldsymbol{x}),\dots, {f}^d(\boldsymbol{x})]^\top: \mathbb{X} \to \mathbb{R}^{d}$ within a distributed system referring as a MAS consist of $n \in \mathbb{N}$ agents equipped with its own GP model and distinct datasets, where $\boldsymbol{x} = [{x}^1,\dots, {x}^m]^\top \in \mathbb{X} \subset \mathbb{R}^{m}$ with $d,m \in \mathbb{N}$ is considered the agent's state.

\begin{remark}\label{rem_f}
This formulation captures a common setting in distributed systems or MASs~\cite{yangDistributedLearningConsensus2021,ledererCooperativeControlUncertain2023,yang2024cooperative,yuan2024lightweight,dai2024decentralized,dai2024cooperative}, where the function $\boldsymbol{f}(\cdot)$ is fully described or characterized by agents possessing identical dynamics in homogeneous systems or identical unknown functions in heterogeneous systems.
Compared to approaches that rely on local or sparse GP models, our framework emphasizes collaborative learning. Thus, leveraging a communication network, these agents utilize their GP models to collaboratively infer the same unknown function. 
Additionally, the distributed formulation generalizes to ensemble learning settings, where a single system may aggregate multiple local models for improved robustness and accuracy.
\end{remark}

The primary focus of this paper revolves around the development of an estimation function $\hat{\boldsymbol{f}}(\cdot)$ for predicting the unknown function $\boldsymbol{f}(\cdot)$ in a cooperative manner. 
In general, the estimation for the $j$-th $(j=1,\dots,d)$ dimension of the unknown function $\boldsymbol{f}(\cdot)$ of the agent $i$  for any query point $\boldsymbol{x} \in \mathbb{X}$ is defined in an aggregation fashion as
\begin{equation}\label{eq_tilde_h}
    \hat{f}_i^j(\boldsymbol{x}) = \sum_{s=1}^n w_{is}^j \mu_s^j(\boldsymbol{x}),
\end{equation}
where $\mu_s^j(\boldsymbol{x}_i)$ is the estimation function obtained by the individual GP model illustrated in the following \cref{subsec_GPR} and $w_{is}^j$ is the aggregation weight, which is calculated by the proposed EIGP learning framework detailed in \cref{sec_EIGP}.

\subsection{Gaussian Process Regression}
\subsubsection{Training Dataset} Capitalizing on the advantages inherent to distributed systems, the training dataset is able to be partitioned and stored in individual agents with distinct subsets of data. As the data are observed and measured in a discrete-time manner, these individual datasets at time $t_k\in \mathbb{R}_+$, $k\in\mathbb{N}$, are denoted as $\mathbb{D}_i(t_k) = \{ \boldsymbol{X}_i(t_k), \boldsymbol{Y}_i(t_k) \}$, where each dataset comprises a collection of $N_i(t_k) \in \mathbb{N}$ data pairs, with each data pair represented as $\{(\boldsymbol{x}_{i,k}, \boldsymbol{y}_{i,k})\}$, where $i\in\mathcal{V}$, $k= 1,\dots, N_i(t_k)$. To elaborate further, we define $\boldsymbol{X}_i(t_k)$ as the set containing input data samples, specifically $\boldsymbol{X}_i(t_k) = \{\boldsymbol{x}_{i,1},\dots, \boldsymbol{x}_{i, N_i(t_k)} \}$ with $\boldsymbol{x}_{i,k} = [{x}_{i,k}^j]_{j=1,\dots,m}$, and $\boldsymbol{Y}_i(t_k)$ as the set encompassing the corresponding output data samples, namely $\boldsymbol{Y}_i(t_k) = \{\boldsymbol{y}_{i,1},\dots, \boldsymbol{y}_{i, N_i(t_k)} \}$ with $\boldsymbol{y}_{i,k} = [{y}_{i,k}^j]_{j=1,\dots,d}$. Therefore, this distributed data storage arrangement effectively leverages the parallelism and localized data processing capabilities of the agents, where the subsets of training data satisfy the following assumption.	
\begin{assumption} \label{ass_dataset}
    For every data pair $\{(\boldsymbol{x}_{i,k}, \boldsymbol{y}_{i,k})\}$ associated with agent $i \in \mathcal{V}$ in the MAS at any given time instance, where $k\in \mathbb{N}$, it holds $y_{i,k}^j = f_i^j(\boldsymbol{x}_{i,k})  + \omega_{i,k}$ for each dimension $j=1,\dots, d$. 
    The term $\omega_{i,k}$ refers to the measurement noise, which follows a zero-mean, independent, and identical Gaussian distribution with variance  $\sigma_{\omega_i}^2$, where $\sigma_{\omega_i} \in \mathbb{R}_+$.
\end{assumption}
Assumption \ref{ass_dataset} shows that each agent possesses the capability to independently collect data and offers the prospect of continuously improving GP models through ongoing online data collection. This feature highlights the adaptability and potential for refinement in the modeling process, as it allows for the incorporation of new data to enhance the accuracy of the learned dynamics over time. 

\subsubsection{Online Learning with GPs}
\label{subsec_GPR}
In order to model the multi-dimensional unknown function $\boldsymbol{f}(\cdot)$, Gaussian process regression is employed. Moreover, the estimation of the $j$-th dimension of $\boldsymbol{f}(\cdot)$ is modeled as a distribution of $f^j(\cdot)$ by the prior mean $m(\cdot): \mathbb{X} \to \mathbb{R}$ and the kernel function $\kappa(\cdot,\cdot): \mathbb{X} \times \mathbb{X} \to \mathbb{R}_{0,+}$. The prior mean function, typically used to capture the known aspects of the function $f(\cdot)$, is set to $m(\cdot) = 0$ for generalization in the modeling process. While the kernel function $\kappa(\cdot)$ calculates the covariance between two training inputs and satisfies the following assumption.
\begin{assumption} \label{ass_GP}
    The continuous scalar function $f(\cdot)$ is a sample from a Gaussian process with a Lipschitz kernel $\kappa(\|\boldsymbol{x} - \boldsymbol{x}'\|) = \kappa(\boldsymbol{x}, \boldsymbol{x}')$ w.r.t. $\|\boldsymbol{x} - \boldsymbol{x}'\|$.
\end{assumption}
The conditions on the kernel $\kappa(\cdot)$ can be met by selecting any continuous function that represents the continuity of $f(\cdot)$, and subsequently, the Lipschitz continuity is established by taking into account the bounded input domain $\mathbb{X}$. 
It is noteworthy that many widely used kernels, such as the square exponential kernel, inherently fulfill the Lipschitz continuity condition~\cite{rasmussenGaussianProcessesMachine2006}, which is written as 
\begin{equation}
    \kappa(\boldsymbol{x},\boldsymbol{x}' )=\sigma_{\kappa}^{2} \exp (-{\|\boldsymbol{x}-\boldsymbol{x}'\|^{2}}/{2 \ell_{\kappa}^{2}}),
\end{equation}
where $\boldsymbol{x},\boldsymbol{x}' \in \mathbb{X}$, $\sigma_{\kappa} \in \mathbb{R}_+$ and $\ell_{\kappa} \in \mathbb{R}_+$ are the hyper-parameters.
By leveraging a pre-defined kernel function, the prediction of the unknown function $\boldsymbol{f}(\cdot)$ becomes feasible through a GP model equipped within each agent, utilizing its respective training dataset. We consider the data set at $t_k$ denoted $\mathbb{D}_i(t_k)$ with $N_i(t_k)$ for the agent $i\in \mathcal{V}$ satisfying Assumption \ref{ass_dataset}, then the $j$-th dimensional conditional posterior mean and the variance of the agent $i$ at any query point $\boldsymbol{x} \in \mathbb{X}$ are
\begin{align} \label{eqn_GP_prediction}
    &\mu_i^j(\boldsymbol{x}|\mathbb{D}_i(t_k)) = \boldsymbol{k}(\boldsymbol{x}, \boldsymbol{X}_i(t_k))^\top \bar{\boldsymbol{K}}(\boldsymbol{X}_i(t_k))^{-1} \boldsymbol{y}_i^j(t_k), \\
\label{eqn_GP_variance}    &{\sigma_i^{j}}^{2}(\boldsymbol{x}|\mathbb{D}_i(t_k)) = \kappa(0) \\
    &\qquad \qquad \qquad- \boldsymbol{k}(\boldsymbol{x}, \boldsymbol{X}_i(t_k))^{\!\top} \bar{\boldsymbol{K}}(\boldsymbol{X}_i(t_k))^{-1} \boldsymbol{k}(\boldsymbol{x}, \boldsymbol{X}_i(t_k)), \nonumber
\end{align}
where $\bar{\boldsymbol{K}}(\boldsymbol{X}_i(t_k)) = \boldsymbol{K}(\boldsymbol{X}_i(t_k) ) + \sigma_{\omega}^2 \boldsymbol{I}_{N_i(t_k)}$, the Gram matrix denotes
$\boldsymbol{K}(\boldsymbol{X}_i(t_k)) =[\kappa(\boldsymbol{x}_{i,p}, \boldsymbol{x}_{i,q})]_{p,q = 1,\dots, N_i(t_k)}$. The vector $\boldsymbol{k}(\boldsymbol{x}, \boldsymbol{X}_i(t_k)) =[\kappa(\boldsymbol{x}, \boldsymbol{x}_{i,1}),\dots, \kappa(\boldsymbol{x}, \boldsymbol{x}_{i,N_i(t_k)})]^\top$ and $\boldsymbol{y}_i^j(t_k) = [y_{i,1}^j,\dots, y_{i,N_i(t_k)}^j]^\top$. 
Note that \( \kappa(0) \) represents the kernel function evaluated at a single input, as defined in \cref{ass_GP}. Consequently, \( \kappa(0) = \kappa(\boldsymbol{x}, \boldsymbol{x}) \). 
If all the calculations of the posterior mean and variance at any query points are under the latest dataset, computational complexity of the matrix inversion in \cref{eqn_GP_prediction} is  $\mathcal{O}(N_i(t_k)^3)$ and computational complexity of \cref{eqn_GP_variance}  is $\mathcal{O}(N_i(t_k))$, respectively. 
We drop it for clarity of presentation, i.e. $\mu_i^j(\boldsymbol{x}|\mathbb{D}_i(t_k)) = \mu_i^j(\boldsymbol{x})$, ${\sigma_i^{j}}^{2}(\boldsymbol{x}|\mathbb{D}_i(t_k))={\sigma_i^{j}}^{2}(\boldsymbol{x})$, indicating the computation as consistently reliant upon the real-time data set.

\section{Error-Informed Cooperative Learning} \label{sec_EIGP}
\subsection{Error-Informed Prediction Quantification} 
To quantify the predictive performance of the GP model embedded within each agent of the MAS, we introduce an error-informed metric. As an agent accumulates sequential training data, it becomes imperative to assess the accuracy of predictions made by the agent based on its continually updated dataset. For each prediction executed by the agent $i$, its precision is evaluated through the computation of a prediction error for the $j$-th prediction at a specific time instance $t_k$. Hence, we define the prediction error as
\begin{equation}\label{eq_predictionError}
e_{i}^j(\boldsymbol{x}_{i,p}, t_k) \!= \! \mu_i^j(\boldsymbol{x}_{i,p})  \!- \! y_{i,p}^j(t_k), ~ \forall \boldsymbol{x}_{i,p} \in \boldsymbol{X}_i(t_k),
\end{equation}
 where $y_{i,p}^j$ is the $j$-th dimension of $\boldsymbol{y}_{i,p}$ associated with its the training input $\boldsymbol{x}_{i,p}$.
Furthermore, we define the prediction error set $\mathbb{E}_i(t_k)=\{\boldsymbol{e}_i^j(t_k)\}_{j=1,\dots,d}$, where $\boldsymbol{e}_i^j(t_k) = [{e}_i^j(\boldsymbol{x}_{i,p},t_k)]_{\boldsymbol{x}_{i,p}\in\boldsymbol{X}_i(t_k)}$. 
Notably, this prediction error term serves as a component of the posterior mean of GPR, which is shown in the following property.
\begin{property}
\label{prop_mu}
For a given training dataset $\mathbb{D}_i(t_k)$, containing $N_i(t_k)$ data pairs that satisfy \cref{ass_dataset}, the prediction for the $j$-th dimension of the function $f_i^j(\cdot)$, evaluated at any point $\bm{x} \in \mathbb{X}$ at time $t_k$, is obtained using Gaussian Process (GP) regression following \cref{ass_GP}. 
The prediction is derived through an error-informed weighted combination, where each kernel value $\kappa(\bm{x}, \bm{x}_{i,p})$ is scaled by the prediction error term $e_{i}^j(\boldsymbol{x}_{i,p}, t_k)$ with the data pairs, $p = 1, \dots, N_i(t_k)$.
More specifically, the expression for the mean prediction $\mu_i^j(\bm{x})$ can be reformulated as
\begin{equation}
\label{eqn_GP_prediction_e}
    \mu_i^j(\bm{x}) = - \sigma_{\omega}^{-2} \sum_{p=1}^{N_i(t_k)} e_{i}^j(\boldsymbol{x}_{i,p}, t_k) \kappa(\bm{x}, \bm{x}_{i,p}).
\end{equation}
\end{property}
\begin{IEEEproof}
{
To facilitate the proof, we introduce two concatenated variables $\bm{e}_i^j(t_k)$ and $\bm{\mu}_i^j(t_k)$, which represent the evolving vector values with respect to the current time $t_k$ as follows
\begin{align*}
    \bm{e}_i^j(t_k) &= \big[ e_i^j(\boldsymbol{x}_{i,1}, t_k), \dots, e_i^j(\boldsymbol{x}_{i,N_i(t_k)}, t_k) \big]^\top , \\
    \bm{\mu}_i^j(t_k) &= \big[ \mu_i^j(\bm{x}_{i,1}), \dots, \mu_i^j(\bm{x}_{i,N_i(t_k)}) \big]^\top.
\end{align*}
According to the definition of the prediction error in \cref{eq_predictionError} and the posterior mean function in \cref{eqn_GP_prediction}, the concatenated prediction error is expressed as
\begin{align}
\label{eq_e_i_j}
    \bm{e}_i^j(t_k) &= \bm{\mu}_i^j(t_k) - \bm{y}_i^j(t_k) \nonumber\\
    &= \big[ \bm{k}(\bm{x}_{i,1}, \bm{X}_i(t_k)), \dots, \bm{k}(\bm{x}_{i,N_i(t_k)}, \bm{X}_i(t_k)) \big]^\top  \nonumber \\
    &\quad \times \bar{\boldsymbol{K}}_i(\boldsymbol{X}_i(t_k))^{-1} \bm{y}_i^j(t_k) - \bm{y}_i^j(t_k).
\end{align}
Given the definition of the Gram matrix $\boldsymbol{K}(\boldsymbol{X}_i(t_k))$, it is straightforward to see that
\begin{align}
\label{eq_e_i_j2}
    \bm{e}_i^j(t_k) &= \big[ \bm{K}_i(\bm{X}_i(t_k)) \bar{\boldsymbol{K}}_i(\bm{X}_i(t_k))^{-1} - \bm{I}_{N_i(t_k)} \big] \bm{y}_i^j(t_k) \nonumber\\
    &= - \sigma_{\omega}^2 \bar{\boldsymbol{K}}_i(\boldsymbol{X}_i(t_k))^{-1} \bm{y}_i^j(t_k).
\end{align}
Therefore, considering \cref{eq_e_i_j2}, the posterior mean function is reformulated as
\begin{align*}
    \mu_i^j(\bm{x}) &= - \sigma_{\omega}^{-2} \boldsymbol{k}(\bm{x}, \boldsymbol{X}_i(t_k))^\top \bm{e}_i^j(t_k),
\end{align*}
which is equivalent to the expression in \eqref{eqn_GP_prediction_e}.
}
\end{IEEEproof}
This property offers an alternative method for computing GP predictions, incorporating the defined prediction errors. Additionally, it provides a systematic approach for optimizing the update of prediction errors after the GP model update. 

To effectively select the most relevant data points and evaluate the accuracy of inferring the unknown function within GP models as the dataset expands in real-time, we leverage the kernel function of GP, which characterizes the correlations between data points based on their distances or similarities. For the $i$-th agent at time $t_k$, we define an index set as follows
\begin{equation}\label{eq_indexSet}
    \mathbb{I}_i(\boldsymbol{x}(t_k),\rho) = \{ p\in \{1, \cdots, N_i(t_k)\} \,|\, \kappa(\bm{x}_{i,p}, \bm{x}(t_k)) \ge \rho\},
\end{equation}
where the selection variable $0\leq \rho \leq \kappa(0)$ sets the lower bound for the elements of $\boldsymbol{k}(\boldsymbol{x}(t_k), \boldsymbol{X}_i(t_k))$. This construction enables the selection of data points based on a similarity threshold $\rho$.
Moreover, the threshold factor $\rho$ can be set as a constant value or a dynamic value, such as the mean, median, and minimum value of the elements of $\boldsymbol{k}(\boldsymbol{x}(t_k), \boldsymbol{X}_i(t_k))$.
Moreover, we define the complement of $\mathbb{I}_i(\boldsymbol{x}(t_k),\rho)$ as $\bar{\mathbb{I}}_i(\boldsymbol{x}(t_k),\rho)$, where $\mathbb{I}_i(\boldsymbol{x}(t_k),\rho) \cup \bar{\mathbb{I}}_i(\boldsymbol{x}(t_k),\rho) = \{1, \cdots, N_i(t_k)\}$ and $\mathbb{I}_i(\boldsymbol{x}(t_k),\rho) \cap \bar{\mathbb{I}}_i(\boldsymbol{x}(t_k),\rho) = \emptyset$. 
Then, utilizing the defined selected index set $\mathbb{I}_i$ and the reformulated posterior prediction in \eqref{eqn_GP_prediction_e}, we construct a distance-aware prediction error variable $\varepsilon_i(\cdot)$ to evaluate the quality of the GP models, which takes into account the spatial relationships between data points and prediction errors. 
Therefore, we design this error-informed distance-aware prediction evaluation term $\varepsilon_i(\boldsymbol{x}(t_k))$ for agent $i$ denoted by
\begin{align}\label{eq_distanceAwareError}
    \varepsilon_i(\boldsymbol{x}(t_k)) = \frac{\big\| \sum_{p \in \mathbb{I}_i(\boldsymbol{x}(t_k),\rho)} \kappa(\bm{x}_{i,p}, \bm{x}(t_k)) \bm{e}_{i}(\boldsymbol{x}_{i,p}, t_k) \big\|}{\lambda \rho |\bar{\mathbb{I}}_i(\boldsymbol{x}(t_k),\rho)| }, \!
\end{align}
where $\bm{e}_{i}(\cdot) = [{e}_{i}^1(\cdot),\dots, {e}_{i}^d(\cdot)]^\top$,  and 
\begin{align}
    &\label{eq_lambda}\lambda = 2 \sqrt{d \beta_{\delta} \kappa(0)} + \sigma_{\omega} \big(2 \sqrt{d\log (1 / \delta_n)} + 2 \log \delta_n^{-1} + d \big)^{1/2}, \\
    &\label{eq_beta_delta}\beta_{\delta} = 2 \sum_{j=1}^{m} \log{\Big( \frac{\sqrt{m}} {2 \tau} (\bar{x}^{j} - \underline{x}^{j}) + 1 \Big)} - 2\log{\delta},
\end{align}
for the selected $\tau \in \mathbb{R}_+$ and $\delta, \delta_n \in (0,1)$. The scalar variables $\bar{x}^{j}$ and $\underline{x}^{j}$ represent the maximum and minimum values of the $j$-th dimension of the elements in $\mathbb{X}$, respectively. 
\begin{remark}
The variable $\lambda$ represents the quantified accuracy loss of the approximated prediction. 
The parameter $\beta_{\delta}$ is derived using the uniform error bound from \cite{ledererCooperativeControlUncertain2023}, and further refined through the tighter bound in \cite{dai2023can} to quantify the prediction error. 
Both $\lambda$ and $\beta_{\delta}$ will be clarified in the \eqref{eq_mu_tildemu} and \eqref{eq_mu_f} in the proof of \cref{lemma_prediction_loss}, respectively.
\end{remark}

This qualitative term $\varepsilon_i$ encompasses prediction errors that are influenced by data points within a spatial distance of $\rho$ from the current prediction location $\boldsymbol{x}(t_k)$ within the dataset of agent $i$. 
Moreover, we can utilize it to evaluate the upper bound of the proportion of accuracy loss between the full posterior mean $\boldsymbol{\mu}_i(\boldsymbol{x}) = [{\mu}_i^1(\boldsymbol{x}), \dots, {\mu}_i^d(\boldsymbol{x})]^\top$ and the approximated prediction defined by
\begin{align}\label{eq_tilde_mu}
    \tilde{\boldsymbol{\mu}}_i(\boldsymbol{x}(t_k)) &= [\tilde{\mu}_i^1(\boldsymbol{x}(t_k)), \dots, \tilde{\mu}_i^m(\boldsymbol{x}(t_k))]^\top \\
    &=- \sigma_{\omega}^{-2} \sum_{p \in \mathbb{I}_i(\boldsymbol{x}(t_k))} \kappa(\bm{x}(t_k), \bm{x}_{i,p}) \bm{e}_{i}(\boldsymbol{x}_{i,p}, t_k),\nonumber
\end{align}
respective to itself, which is illustrated as follows.

\begin{lemma} \label{lemma_prediction_loss}
    Consider a GP model with data set $\mathbb{D}_i(t_k)$ available at $t_k$ satisfying \cref{ass_dataset} and \ref{ass_GP}.
    Choose $\tau \in \mathbb{R}_+$, $\delta \in (0,1)$ and $\delta_n \in (0,1)$ such that $\delta_{\rho} = 1 - n + n(1 - \delta)^d - \delta_n$ belongs to $(0,1)$, then $\varepsilon_i(\boldsymbol{x}(t_k))$ represents the upper bound of the relative accuracy loss in the computation of the posterior mean $\bm{\mu}_i(\bm{x}(t_k))$, which is represented as
    \begin{align} \label{eqn_loss_rho}
    \frac{ \| \bm{\mu}_i(\bm{x}(t_k)) - \tilde{\bm{\mu}}_{i}(\bm{x}(t_k)) \| }{\| \tilde{\bm{\mu}}_{i}(\bm{x}(t_k)) \|} \le \frac{1}{\varepsilon_i(\boldsymbol{x}(t_k))} 
    \end{align}
    with probability of at least $1 - \delta_{\rho}$.
\end{lemma}
\begin{IEEEproof}
Employing the result in \cref{prop_mu}, it is straightforward to derive the bound of relative accuracy loss, which is written as
\begin{align} \label{eqn_loss_outer_rho}
    &\| \bm{\mu}_i(\bm{x}(t_k)) - \tilde{\bm{\mu}}_{i}(\bm{x}(t_k)) \| \nonumber\\
    &= \sigma_{\omega}^{-2} \Big\| \sum_{p \in \bar{\mathbb{I}}_i(t_k,\rho)} \kappa(\bm{x}(t_k), \bm{x}_{i,p})  \bm{e}_{i}(\boldsymbol{x}_{i,p}, t_k) \Big\| \nonumber \\
    &= \sigma_{\omega}^{-2} \Big\| \big([\kappa(\bm{x}(t_k), \bm{x}_{i,p_1}), \cdots, \kappa(\bm{x}(t_k), \bm{x}_{i,p_{|\bar{\mathbb{I}}_i(t_k,\rho)|}})] \otimes \bm{I}_m \big) \nonumber \\
    &\quad \times \big[\bm{e}_{i}^\top(\boldsymbol{x}_{i,p_1}, t_k), \cdots, \bm{e}_{i}^\top(\boldsymbol{x}_{i,p_{|\bar{\mathbb{I}}_i(t_k,\rho)|}}, t_k)\big]^\top \Big\| \nonumber \\
    &\le \sigma_{\omega}^{-2} \rho \sqrt{| \bar{\mathbb{I}}_i(t_k,\rho) |} \Big(\big\| \bm{f}(\boldsymbol{X}_{i,\bar{\mathbb{I}}_i(t_k,\rho)}) - \bm{y}_{i,\bar{\mathbb{I}}_i(t_k,\rho)}\big\| \nonumber \\
    &\quad + \big\| \bm{\mu}_i(\boldsymbol{X}_{i,\bar{\mathbb{I}}_i(t_k,\rho)}) - \bm{f}(\boldsymbol{X}_{i,\bar{\mathbb{I}}_i(t_k,\rho)})\big\| \Big), 
\end{align}
where $p_1, \cdots, p_{\bar{\mathbb{I}}_i(t_k,\rho)} \in \bar{\mathbb{I}}_i(t_k,\rho)$ represents the indices and $\boldsymbol{X}_{i,\bar{\mathbb{I}}_i(t_k,\rho)} = \{ \boldsymbol{x}\in \boldsymbol{X}(t_k)| \kappa(\bm{x}, \bm{x}_i(t_k)) < \rho \}$. The concatenated approximated prediction, true value of $\bm{f}(\cdot)$ and measurement are denoted as 
\begin{align}
    &\bm{\mu}_i(\boldsymbol{X}_{i,\bar{\mathbb{I}}_i(t_k,\rho)}) =[\bm{\mu}_i^\top(\boldsymbol{x}_{i,p_1}), \dots , \bm{\mu}_i^\top(\boldsymbol{x}_{i,p_{|\bar{\mathbb{I}}_i(t_k,\rho)|}})]^\top \\
    &\bm{f}(\boldsymbol{X}_{i,\bar{\mathbb{I}}_i(t_k,\rho)}) \!=\! [\bm{f}^\top(\bm{x}_{i,p_1}), \!\cdots\!, \bm{f}^\top(\bm{x}_{i,p_{|\bar{\mathbb{I}}_i(t_k,\rho)|}})]^\top\\
    &\bm{y}_{i,\bar{\mathbb{I}}_i(t_k,\rho)} = [\bm{y}_{i,p_1}^\top, \cdots, \bm{y}_{i,p_{|\bar{\mathbb{I}}_i(t_k,\rho)|}}^\top]^\top,
\end{align}
respectively. 
According to the definition of $\beta_{\delta}$ in \eqref{eq_beta_delta}, the prediction error $\bm{\mu}_i(\boldsymbol{x}_{i,p}) - \bm{f}(\bm{x}_{i,p_1})$ is bounded as
\begin{align}\label{eq_mu_f}
    \| \bm{\mu}_i(\boldsymbol{x}_{i,p}) - \bm{f}(\bm{x}_{i,p_1}) \| &\le 2 \sqrt{d \beta_{\delta}} \sigma_i(\boldsymbol{x}_{i,p}) \le 2 \sqrt{d \beta_{\delta} \kappa(0)}
\end{align}
with a probability of at least $(1 - \delta)^d$ due to the independence of each dimension in $\bm{f}(\cdot)$ \cite{yangDistributedLearningConsensus2021,yang2024cooperative}.
Next, considering the union bound, the concatenated approximated prediction error bound is written as
\begin{align} \label{eqn_concatenated_prediction_error_outer_rho}
\| \bm{\mu}_i(\boldsymbol{X}_{i,\bar{\mathbb{I}}_i(t_k,\rho)}) \!-\! \bm{f}(\boldsymbol{X}_{i,\bar{\mathbb{I}}_i(t_k,\rho)})\| \le 2 \sqrt{d \beta_{\delta} \kappa(0) |\bar{\mathbb{I}}_i(t_k,\rho)|} 
\end{align}
with a probability of at least $\delta_{\rho} \le 1 - |\bar{\mathbb{I}}_i(t_k,\rho)| (1 - (1-\delta)^d)$. Moreover, considering the assumption of the distribution for the measurement noise in Assumption \ref{ass_dataset}, $ \bm{f}(\boldsymbol{X}_{i,\bar{\mathbb{I}}_i(t_k,\rho)}) - \bm{y}_{i,\bar{\mathbb{I}}_i(t_k,\rho)}$ is a vector of $d |\bar{\mathbb{I}}_i(t_k,\rho)|$ independent, identical, zero-mean Gaussian variables with variance $\sigma_{\omega}^2$, which results in $\chi$-distribution for $\sigma_{\omega}^{-2} \|  \bm{f}(\boldsymbol{X}_{i,\bar{\mathbb{I}}_i(t_k,\rho)}) - \bm{y}_{i,\bar{\mathbb{I}}_i(t_k,\rho)} \|^2 $ with degrees of freedom $d |\bar{\mathbb{I}}_i(t_k,\rho)|$. Then, using the result in \cite{laurent2000adaptive}, the concatenated measurement noise is bounded with a probability of at least $1 - \delta_n$ as
\begin{align} \label{eqn_concatenated_measurement_noise_outer_rho}
    &\|  \bm{f}(\boldsymbol{X}_{i,\bar{\mathbb{I}}_i(t_k,\rho)}) - \bm{y}_{i,\bar{\mathbb{I}}_i(t_k,\rho)} \|^2  \nonumber \\
    &\le \Big(2 \sqrt{d |\bar{\mathbb{I}}_i(t_k,\rho)| \log \delta_n^{-1}} + 2 \log \delta_n^{-1} + d |\bar{\mathbb{I}}_i(t_k,\rho)| \Big) \sigma_{\omega}^2 \nonumber \\
    &\le \Big(2 \sqrt{d \log \delta_n^{-1}} + 2 \log \delta_n^{-1} + d \Big) |\bar{\mathbb{I}}_i(t_k,\rho)| \sigma_{\omega}^2.
\end{align}
Applying the boundness of concatenated prediction error in \eqref{eqn_concatenated_prediction_error_outer_rho} and measurement noise \eqref{eqn_concatenated_measurement_noise_outer_rho} into \eqref{eqn_loss_outer_rho} and considering their dependency as shown in \eqref{eqn_GP_prediction}. 
According to the definition of $\lambda$ in \eqref{eq_lambda}, the difference in \eqref{eqn_loss_outer_rho} is bounded as
\begin{align}
\label{eq_mu_tildemu}
    \| &\bm{\mu}_i(\bm{x}(t_k)) - \tilde{\bm{\mu}}_{i}(\bm{x}(t_k)) \|\le \lambda \rho  |\bar{\mathbb{I}}_i(t_k,\rho)| \sigma_{\omega}^{-2} \nonumber \\
    &= \frac{\big\| \sigma_{\omega}^{-2} \sum_{p \in \mathbb{I}_i(t_k,\rho)} \kappa(\bm{x}_{i,p}, \bm{x}(t_k)) \bm{e}_{i}(\boldsymbol{x}_{i,p}, t_k) \big\|}{\varepsilon_i(\boldsymbol{x}(t_k))} \nonumber\\
    & = \frac{\| \bm{\mu}_{i,\rho}(\bm{x}(t_k), \mathbb{D}_i(t_k)) \|}{\varepsilon_i(\boldsymbol{x}(t_k))}.
\end{align}
with a probability of at least $1 - \delta_{\rho}$ using the union bound, which concludes the proof.
\end{IEEEproof}
\begin{remark}
    Lemma \ref{lemma_prediction_loss} indicates the designed $\varepsilon_i(\boldsymbol{x}(t_k))$ serves as a metric to assess prediction loss and represents a proportional measure. 
    Consequently, due to the inverse relationship between the proposed error-informed distance-aware prediction evaluation term and the relative accuracy loss in the equation \eqref{eqn_loss_rho}, a higher value $\varepsilon_i(\boldsymbol{x}(t_k))$ indicates smaller prediction loss.
\end{remark}

\subsection{Model Update Strategy}
Effectively managing real-time streaming training data and GP models is essential for ensuring fast prediction, error quantification, and optimal memory utilization in the context of online learning. To elucidate the sequential update process, we focus on the dynamic aspects of dataset maintenance and Gram matrix adaptation, particularly when removing data points. For further handling the data deletion, a well-defined strategy for updating both the dataset and the associated Gram matrix is imperative. By recording the previous data and Gram matrix, we update results sequentially, which are formulated as follows
\begin{align}
 \label{eq_kboder2}&\boldsymbol{k}\big(\boldsymbol{x}_i(t_{k+1}), \boldsymbol{X}_i(t_{k+1})) =  \big[ \bar{\boldsymbol{k}}(\boldsymbol{x}_i(t_{k+1})), \kappa(0) \big]^{\top},\\
 &\bar{\boldsymbol{k}}(\boldsymbol{x}_i(t_{k+1})) = \boldsymbol{k}\big(\boldsymbol{x}_i(t_{k+1}), \boldsymbol{X}_i(t_{k}))^{\top},\\
 \label{eq_Kmatrix2}&\boldsymbol{K}(\boldsymbol{X}_i(t_{k+1}))=  \begin{bmatrix}
 \boldsymbol{K}(\boldsymbol{X}_i(t_{k})) & \bar{\boldsymbol{k}}(\boldsymbol{x}_i(t_{k+1})) \\
\bar{\boldsymbol{k}}(\boldsymbol{x}_i(t_{k+1}))^{\top}&  \kappa(0)
\end{bmatrix},
\end{align}
with the updated data $\boldsymbol{X}_i(t_{k+1}) = \big[\boldsymbol{X}_i(t_{k}) , \boldsymbol{x}_i(t_{k+1})\big]$, $ 
\boldsymbol{Y}_i(t_{k+1}) = \big[\boldsymbol{Y}_i(t_k),  \boldsymbol{y}_i(t_{k+1})\big]$. Having this update procedure, we can deal with the persistent expansion of the training datasets with a data deletion mechanism. 
When the dataset of the $i$-th agent reaches the predefined threshold $\bar{N}_i \in\mathbb{N}$ at the time point $t_k$, the input training data point to be removed is defined as 
\begin{align}\label{eq_delete_x}
    \tilde{\boldsymbol{X}}_i(t_k)  = \argmin_{\boldsymbol{x}\in \boldsymbol{X}_i(t_k)}\{\kappa(\boldsymbol{x},\boldsymbol{x}_i(t_{k+1})) \}.
\end{align}

\begin{remark}
There are multiple ways to update the dataset while balancing efficiency and information retention. 
One straightforward approach is the sliding window method, which removes the oldest data points to maintain dataset continuity. 
In contrast, EIGP employs kernel similarity to prune the dataset, preserving the most relevant data of the state of the dynamical system while ensuring the dataset remains within the storage limits. 
Although sparse GP can selectively retain key data points, thereby losing less information compared to the sliding window and EIGP methods, it requires reconstructing the prediction dataset through computationally intensive optimization processes, making it unsuitable for real-time online learning systems.v
\end{remark}

Furthermore, the corresponding training output set is defined as $\tilde{\boldsymbol{Y}}(t_k) $. Subsequently, the data sets after deletion are reconstructed by 
\begin{subequations}
\label{eq_datasetUpadate}
   \begin{align}
       \boldsymbol{X}_i(t_{k+1}) &= \boldsymbol{X}_i(t_k) \backslash \tilde{\boldsymbol{X}}_i(t_k), \\ \boldsymbol{Y}_i(t_{k+1}) &=\boldsymbol{Y}_i(t_k) \backslash \tilde{\boldsymbol{Y}}_i(t_k).
\end{align} 
\end{subequations}
This iterative process ensures that the agent's dataset remains within the defined threshold preventing inflation of the dataset size, where the computational complexity in \cref{eq_delete_x} is  \(O(N_i(t_k))\) using the linear scan approach. 
Concurrently, the Gram matrix necessitates reallocation as elucidated in the \cref{alg_errorUpdate} to prevent complete recalculation. 
Moreover, updating the prediction error is facilitated through the reformulated posterior prediction that eschews redundant dual computations. 
Specifically, \cref{eq_e_i_j} demonstrates that updating the prediction error term \(\boldsymbol{e}_i^j(t_k)\) involves the computation of \(\bar{\boldsymbol{K}}_i(\boldsymbol{X}_i(t_k))^{-1} \boldsymbol{y}_i^j(t_k)\). 
However, this computation can be efficiently retrieved from the cached results, which previously had been calculated by the posterior mean function in \cref{eqn_GP_prediction}. Consequently, updating the prediction error in real-time is feasible without imposing additional computational burdens.

To clarify the proposed model update strategy, an algorithmic representation of the process is provided in pseudocode. This includes both the prediction error update and the data deletion operation, as illustrated in \cref{alg_errorUpdate}.
\begin{algorithm} [t]
\caption{Model and Data Update}
\label{alg_errorUpdate}
\begin{algorithmic} [1]
    \STATE Initialize $\boldsymbol{X}_i(t_0),\boldsymbol{Y}_i(t_0)$ and set $\rho$ function
    \WHILE{$t_k\leq t_f$}
    \STATE Clear $\mathbb{E}_i(t_{k-1})$   
            \STATE Update $\boldsymbol{X}_i(t_k)$ and $\boldsymbol{Y}_i(t_k)$
            \FOR{$j \leq m$}
                \STATE Update $\boldsymbol{e}_i^j(t_k)$ $\gets$ \eqref{eq_e_i_j}
            \ENDFOR
            \STATE Add all $\boldsymbol{e}_i^j(t_k)$ to $\mathbb{E}_i(t_k)$
            \IF{$N_i(t_k) \geq \bar{N}_i$}
                \STATE Find $\tilde{\boldsymbol{X}}_i(t_k)$ and $\tilde{\boldsymbol{Y}}_i(t_k)$ $\gets$ \eqref{eq_delete_x}
                \STATE Reallocate $\boldsymbol{K}(\boldsymbol{X}_i(t_{k+1}))$ by removing the columns and rows associated with $\boldsymbol{x} \in \tilde{\boldsymbol{X}}_i(t_k)$
                \STATE Update $\boldsymbol{X}_i(t_{k+1})$ and $\boldsymbol{Y}_i(t_{k+1})$ $\gets$ \eqref{eq_datasetUpadate}
            \ENDIF      
    \ENDWHILE
\end{algorithmic}
\end{algorithm}

\subsection{Error-informed Gaussian Process Regression}\label{subsec_EIGP}
To enhance the efficiency of the cooperative learning process, the distributed learning framework must extend its focus beyond mere subsystem quantity to encompass considerations of quality. 
Consequently, two algorithms based on EIGP are devised to enable more accurate and efficient predictions, both of which can be integrated within the following framework
\begin{equation}\label{eq_EIGP_framework}
    \hat{f}_i^j(\boldsymbol{x}) = \sum_{s\in \mathcal{S}_i(\boldsymbol{x})} w_{is}^j(\boldsymbol{x}) \tilde{\mu}_s^j(\boldsymbol{x}), ~\forall i\in \mathcal{V}.
\end{equation}
Here, we drop the time $t_k$ in $\boldsymbol{x}(t_k)$ for notation simplicity.
The computational complexity associated with computing the $i$-th agent's posterior mean ${\mu}_i^j(\boldsymbol{x})$ employing all available data is $\mathcal{O}(N_i^2)$. 
However, utilizing the fast prediction in \cref{eq_tilde_mu} reduces the computational cost to $\mathcal{O}(|\mathbb{I}_i(\boldsymbol{x})|^2) < \mathcal{O}(N_i^2)$. 
Additionally, the agent set $\mathcal{S}_i(\boldsymbol{x}) \subseteq \bar{\mathcal{N}}_i$ is determined through an error-informed selective function $\phi(\boldsymbol{x}): \mathbb{R}\to  \{0,1\}$. 
This function reflects reduced aggregate predictions in cases where $|\mathcal{S}_i(\boldsymbol{x})|<|\bar{\mathcal{N}}_i|$, illustrating a key feature of EIGP, i.e., the use of $\mathcal{S}_i(\boldsymbol{x})$ to select high-quality neighbors. 
Consequently, the overall computational complexity of the MAS for one prediction is  $\mathcal{O}(|\mathcal{S}_i(\boldsymbol{x})||\mathbb{I}_i(\boldsymbol{x})|^2) < \mathcal{O}(|\bar{\mathcal{N}}_i| N_i^2)$. 
Further elaboration on EIGP is provided in subsequent subsections.

\subsubsection{Greedy EIGP}
In pursuit of identifying the most reliable prediction under the proposed criteria, we present an algorithm with {g}reedy protocol utilizing the {EIGP} (gEIGP). This ensures the selection of a single agent from its neighbors for prediction, which accelerates the prediction process by obviating the engagement of additional computational agents, thereby facilitating a rapid and resource-efficient computational framework. 

Owing to the proposed quantifiable error-informed term $\varepsilon$ in \cref{eq_distanceAwareError}, each agent is able to assess the reliability of its neighbors. 
The error-informed selective function for gEIGP is formulated as
\begin{align}\label{eq_phi_gEIGP}
     \phi_{is}^{g}(\boldsymbol{x}) =  \begin{cases}
        1, & \varepsilon_{s}(\boldsymbol{x}) = \bar{\varepsilon}_i(\boldsymbol{x}) \\
        0, & \text{ otherwise }
    \end{cases}, 
\end{align}
where $\bar{\varepsilon}_i(\boldsymbol{x}) = \max \big( \{ \varepsilon_{s}(\boldsymbol{x}) \}_{s\in \bar{\mathcal{N}}_i} \big )$. 
Since we only pick the optimal one for inferring the unknown function with the largest value $\varepsilon$ for the agent $i$ among its neighbor set $\bar{\mathcal{N}}_i$, we can directly obtain the aggregation weight for all dimension predictions by letting 
\begin{align}\label{eqn_aggregation_weights_gEIGP}
    w_{is}^j(\boldsymbol{x}) = \phi_{is}^{g}(\boldsymbol{x}), ~\forall j=1,\dots,m.
\end{align}
Then, we define the set $\mathcal{S}_i(\boldsymbol{x}) = \{s | \phi_{is}(\boldsymbol{x}) =1 \}$, where binary aggregation elements are utilized to indicate the active or inactive status of agent $s$ in the prediction.
It is noteworthy that, to mitigate the risk of overfitting, where the chosen agent possesses a relatively small dataset, certain strategies can be employed. One straightforward approach is augmenting the volume of data available to the agents. 
Additionally, an alternative strategy is to broaden the engagement with available neighbors without adhering strictly to a greedy selection. 
However, instead of indiscriminately requiring joint predictions from all neighbors, we propose a practical approach for enhanced feasibility in the subsequent subsection.

\subsubsection{Adaptive EIGP}
Despite the inherent diversity in datasets across agents, certain agents exhibit comparable prediction performance metrics, such as prediction accuracy and data correlation. In light of this observation, we introduce a novel approach integrating an {a}daptive strategy with {EIGP} (aEIGP). This approach enables agents to selectively choose neighbors within a specified confidence.

To implement this strategy, we first transform the distance-aware error-informed evaluation term $\varepsilon$ into a Gaussian distribution through the expression 
\begin{align}\label{eq_hat_varepsilon}
     \tilde{\varepsilon}_{is}(\boldsymbol{x}) = 
        \frac{\exp\big({-(\bar{\varepsilon}_i(\boldsymbol{x}) - {\varepsilon}_s(\boldsymbol{x}))^2}/ {2\sigma_{\varepsilon,i}^2(\boldsymbol{x})}\big)}{\sigma_{\varepsilon,i}(\boldsymbol{x})\sqrt{2\pi}}, 
\end{align}
where $\sigma_{\varepsilon,i}(\boldsymbol{x})$ is defined as the standard deviation of the set $\{ \varepsilon_{s}(\boldsymbol{x}) \}_{s\in \bar{\mathcal{N}}_i}$.
This adaptation aims to provide the aEIGP with a mechanism for collaborator selection for cooperative learning, enhancing the protocol's flexibility in accommodating agents with comparable prediction capabilities within a defined confidence range. 
Thus, a selective function is devised based on the characteristics of the Gaussian distribution, which is defined as
\begin{align}\label{eq_phi_aEIGP}
     \phi_{is}^{a}(\boldsymbol{x}) =  \begin{cases}
        1, & {\varepsilon}_{s}(\boldsymbol{x}) \geq \bar{\varepsilon}_{i}(\boldsymbol{x})- \theta\sigma_{\varepsilon,i}(\boldsymbol{x}) \\
        0, & \text{ otherwise }
    \end{cases},
\end{align}
where $\theta$ controls the range of confidence interval. Then, we define the binary vector $\boldsymbol{\phi}_{i}^{a}(\boldsymbol{x}) = [\phi_{is}^{a}(\boldsymbol{x})]_{s \in {\mathcal{S}}_i(\boldsymbol{x}) }$.
For simplicity in notation, we formulate the error-informed aggregation weight vector as 
\begin{align}\label{eq_tildeW}
    \tilde{\boldsymbol{w}}_i(\boldsymbol{x}) 
    = \boldsymbol{\psi}(\boldsymbol{\phi}_{i}^{a}(\boldsymbol{x}) \odot  \tilde{\boldsymbol{\varepsilon}}_{i}(\boldsymbol{x})), ~\forall i \in \mathcal{V},
\end{align}
where the vector $\tilde{\boldsymbol{w}}_i(\boldsymbol{x}) =  [ \tilde{w}_{is}(\boldsymbol{x})]_{s\in{\mathcal{S}}_i(\boldsymbol{x})}$, $\boldsymbol{\phi}_{i}^{a}(\boldsymbol{x}) = [ \phi_{is}^a(\boldsymbol{x}(t_k))]_{s\in{\mathcal{S}}_i(\boldsymbol{x})}$ the symbol $\odot$ denotes the Hadamard product and the vector $\tilde{\boldsymbol{\varepsilon}}_{i}(\boldsymbol{x}) = [\tilde{\varepsilon}_{is}(\boldsymbol{x}(t_k))]_{s\in \bar{\mathcal{N}}_i}$. The vector-valued function $\boldsymbol{\psi}(\cdot):\mathbb{R}^d\to\mathbb{R}^d$, $d\in\mathbb{N}$ with each element as a min-max normalization function ${\psi}_i(\cdot): \mathbb{R}^d \to \mathbb{R}_{0,+}$ is defined as ${\psi}_i(\boldsymbol{v}) = {(v_i- v_{\min})}/{({v}_{\max}-{v}_{\min} )}$, where $\boldsymbol{v}=[v_1,\dots,v_d]$. This function $\boldsymbol{\psi}(\cdot)$ rescales associated with the input vector $\boldsymbol{v}$. The values $v_{\max}$ and $v_{\min}$ are the maximal and minimal elements of the vector $\boldsymbol{v}$, respectively. 

\begin{remark}
As the aggregation weights \cref{eqn_aggregation_weights_gEIGP} and the selective function \cref{eq_phi_aEIGP} can be construed as a comparative process within the set $\{{\varepsilon}_{s}(\boldsymbol{x})\}_{s\in\bar{\mathcal{N}}_i}$, the necessity for computing the value of $\lambda$ in \cref{eq_distanceAwareError} is thereby obviated.
\end{remark}

Since the posterior variance in GPR serves as an indicator of prediction uncertainties, quantifying the confidence degree of predictions with respect to the training dataset \cite{hintonTrainingProductsExperts2002}. 
Employ the similar approaches in the BCM family \cite{trespBayesianCommitteeMachine2000,deisenrothDistributedGaussianProcesses2015}, i.e., integrating posterior variance with the proposed selective aggregation weights involves multiplying an additional factor that considering the prior variance given as $\vartheta_{is}^j(\boldsymbol{x}) = \log((\kappa(0) +{\sigma_{\omega}^{j}}^2)/{\sigma_{is}^{j}}^{2}(\boldsymbol{x})$, we develop the error-informed  aggregation weights for aEIGP\footnote{The generalization of aEIGP and its variations are presented in the \cref{sec_aEIGP_variants}.} as follows
\begin{align}\label{eqn_aggregation_weights_aEIGP}
    &w_{is}^{j}(\boldsymbol{x}) \!=\!  \frac{ \big(\tilde{w}_{is}(\boldsymbol{x})\big)^{\nu} \Big( {\vartheta_{is}^j(\boldsymbol{x}){\sigma_s^{j}}^{-2}(\boldsymbol{x})\tilde{\sigma}{{}_i^{j}}^2(\boldsymbol{x})} \Big)^{(1-\nu)} }  {\Gamma_i(\boldsymbol{x})}, \\
   \label{eqn_aggregation_variances_aEIGP} &\tilde{\sigma}{{}_i^{j}}^{-2}(\boldsymbol{x}) = \sum_{l\in\mathcal{S}_i(\boldsymbol{x})} \tilde{w}_{il}(\boldsymbol{x}) \vartheta_{il}^j(\boldsymbol{x}){\sigma_l^j}^{-2}(\boldsymbol{x}) \\
   &~~~+ \Big(1- \sum_{l\in\mathcal{S}_i(\boldsymbol{x})} (\tilde{w}_{is}(\boldsymbol{x}))^{\nu} ( \vartheta_{is}^j(\boldsymbol{x}))^{(1-\nu)}\Big) / (\kappa(0)  + {\sigma_{\omega}^{j}}^2 ),\nonumber
\end{align}
with 
\begin{align}
    \Gamma_i(\boldsymbol{x}) \!=\!\!\! \sum_{l\in{\mathcal{S}}_i(\boldsymbol{x})}\!\! \big(\tilde{w}_{il}(\boldsymbol{x})\big)^{\nu}\! \Big(\vartheta_{il}^j(\boldsymbol{x}){\sigma_l^{j}}^{-2}(\boldsymbol{x})\tilde{\sigma}{{}_i^{j}}^2\!(\boldsymbol{x}) \Big)^{\!(1-\nu) },\nonumber
\end{align}
where the factor $ \nu \in [0,1] \subset \mathbb{R}$ governs the proportion of influence attributed to the error-informed distributed aggregation weight versus the posterior variance of GPs. Therefore, within the framework of aEIGP, a flexible adjustment is afforded by setting $\nu=0$, which degenerates into a selective POE. Thus, the aggregation weights are solely determined by posterior variance. Conversely, when $\nu=1$, it indicates that the weights rely entirely on error-informed aggregation weights as defined in \cref{eq_tildeW}. In this case, an expedited computation of aggregation weights can be achieved by circumventing variance calculations.

\begin{remark}
It is noteworthy to observe that gEIGP can be regarded as a specific instance of aEIGP when only one selective weight ${w}_{is}^j = 1, ~\forall s \in \bar{\mathcal{N}}_i$. Consequently, aEIGP encompasses a broader spectrum of scenarios. However, gEIGP 
demonstrates an advantage in terms of computational efficiency, as it refrains additional computations, i.e., \cref{eq_hat_varepsilon,eq_tildeW,eqn_aggregation_variances_aEIGP} for obtaining aggregation weights.
\end{remark}

Till here, by substituting either aggregation weights of gEIGP in \cref{eqn_aggregation_weights_gEIGP} or aEIGP in \cref{eqn_aggregation_weights_aEIGP} to \cref{eq_EIGP_framework}, the cooperative joint prediction with the EIGP can be obtained.
To facilitate a better understanding of the algorithm's operation for the $i$-th agent in the MAS, we provide a pseudo-code representation of EIGP in \cref{alg_EIDGP}.
\begin{algorithm} [t]
\caption{EIGP Algorithm}\label{alg_EIDGP}
\begin{algorithmic}[1]
\STATE Choose $mode$ = gEIGP or aEIGP 

\FOR{$s \in \bar{\mathcal{N}}_i$}
    \STATE $\mathbb{I}_s(\boldsymbol{x}(t_k),\rho)$, ${\varepsilon}_{s}(\boldsymbol{x}(t_k))$ ~~$\gets$ \cref{eq_indexSet,eq_distanceAwareError}
    \FOR{$j = 1: m$}
        \SWITCH{$mode$}
        \CASE{aEIGP}
          \STATE $\phi_{is}^g(\boldsymbol{x}(t_k))$, $w_{is}^j(\boldsymbol{x}(t_k))$ $\gets$ \eqref{eq_phi_gEIGP}, \eqref{eqn_aggregation_weights_gEIGP}
        \ENDCASE
        \CASE{gEIGP}
            \STATE $\tilde{\varepsilon}_{is}(\boldsymbol{x}(t_k))$, $\phi_{is}^{a}(\boldsymbol{x}(t_k))$ $\gets$ \eqref{eq_hat_varepsilon}, \eqref{eq_phi_aEIGP}
          \IF{$\nu=1$}
            \STATE $w_{is}^j(\boldsymbol{x}(t_k)) = \tilde{{w}}_{is}(\boldsymbol{x}(t_k))$ $\gets$ \eqref{eq_tildeW}
          \ELSE
          \STATE $w_{is}^j(\boldsymbol{x}(t_k))$ $\gets$  \eqref{eqn_aggregation_weights_aEIGP}
          \ENDIF
        \ENDCASE
      \ENDSWITCH
        
    \ENDFOR
    \STATE  $\hat{f}_i^j(\boldsymbol{x}(t_k))$ $\gets$ \cref{eq_EIGP_framework}
\ENDFOR

\IF{$|\{y_i^j(t_k)\}_{j=1,\dots,d}| = d$}
\STATE Update $\mathbb{D}_i(t_k)$ and $\mathbb{E}_i(t_k)$
\ENDIF

\end{algorithmic}
\end{algorithm}

\subsection{Prediction Guarantee}
In safety-critical MAS applications, it is imperative to quantify the estimation error associated with the unknown function. This quantification ensures that prediction performance maintains an acceptable level of error within a certain region. In this subsection, we conduct an analysis of the joint prediction error bounds of EIGP with real-time datasets by leveraging the inherent probability bounds of GP regression~\cite{srinivasInformationTheoreticRegretBounds2012a,lederer2019uniform}. Therefore, concerning the multi-dimensional prediction guarantee of the individual GP model, we present the following lemma.

\begin{lemma} \label{lemma_prediction_error}
Consider a GP model with data set $\mathbb{D}_i(t_k)$ available at $t_k$ satisfying Assumption \ref{ass_GP} and Assumption \ref{ass_dataset}, respectively.
Choose $\tau \in \mathbb{R}_+$, $\delta \in (0,1)$ such that $\delta_x = 1 - (1 - \delta)^d$ belongs to $(0,1)$, then the prediction error is bounded by
\begin{align} \label{eqn_prediction_error}
    &\| \bm{\mu}_i(\bm{x}(t_k)) - \bm{f}(\bm{x}(t_k)) \| \le \eta_i(\bm{x}(t_k)) \\
    &=  2 \sqrt{d \beta_{\delta} \Big(\kappa(0) - |\mathbb{I}_i(t_k,\rho)| \rho^2\big(|\mathbb{I}_i(t_k,\rho)| \kappa(0) + \sigma_{\omega}^2 \big)^{-1} \Big)},\nonumber 
\end{align}
with probability of at least $1 - \delta_x$.
\end{lemma}
\begin{IEEEproof}
Considering the probabilistic prediction error bound for GP regression in \eqref{eq_mu_f}, it has
\begin{align} \label{eqn_prediction_error_derive1}
	\| \bm{\mu}_i(\bm{x}(t_k)) - \bm{f}(\bm{x}(t_k)) \| \le 2 \sqrt{d \beta_{\delta}} \sigma_{i}(\bm{x}(t_k)), 
\end{align}
with a probability of at least $(1 - \delta)^d$.
Considering the fact that more data set induces smaller posterior variance under positive measurement noise $\sigma_{\omega}^j, \forall j \in \{1, \cdots, d\}$ \cite{wendland2004scattered}, it has $\sigma_{i}(\bm{x}(t_k)) \le \sigma_{i,\rho}(\bm{x}(t_k))$ with $\sigma_{i,\rho}(\bm{x}(t_k))$ as the posterior variance using only $|\mathbb{I}_i(t_k,\rho)|$ with $\kappa(\bm{x}(t_k), \bm{x}_{i,p}) \le \rho$.
Using the property of $\mathbb{I}_i(t_k,\rho)$, the upper bound of $\sigma_{i,\rho}(\bm{x}_i(t_k))$ is written as
\begin{align} \label{eqn_posterior_bound}
    \sigma_{i,\rho}(\bm{x}(t_k)) &\le \kappa(0) - \frac{\| \bm{k}(\bm{x}(t_k), \bm{X}_{i,\mathbb{I}_i(t_k,\rho)})\|^2}  {\underline{\sigma}(\bm{K}_i(\bm{X}_{i,\mathbb{I}_i(t_k,\rho)}(t_k))) + \sigma_{\omega}^2} \nonumber \\
    &\le \kappa(0) - \frac{|\mathbb{I}_i(t_k,\rho)| \rho^2} {|\mathbb{I}_i(t_k,\rho)| \kappa(0) + \sigma_{\omega}^2}, 
\end{align}
where $\underline{\sigma}(\cdot)$ returns the minimal singular value of a matrix, and the second inequality is derived considering the Gershgorin circle theorem. Then, apply the upper bound of $\sigma_{i,\rho}(\bm{x}(t_k))$ in \eqref{eqn_posterior_bound} into \eqref{eqn_prediction_error_derive1}, which concludes the proof.
\end{IEEEproof}

\begin{remark}
    According to the \cref{eqn_prediction_error_derive1} and \eqref{eqn_posterior_bound}, the prediction error bound depends mainly on three key factors: 
    (1) The selection variable $\rho$, which controls the selected data relevance degree.
    (2) The size of set $\mathbb{I}_i(t_k,\rho)$, which reflects the local data density used for prediction.
    (3) The noise variance $\sigma_{\omega}^2$, which quantifies uncertainty in the observed data. 
    Reformulating \cref{eqn_posterior_bound}, we obtain the fact that
    $\sigma_{i,\rho}(\boldsymbol{x}(t_k)) \le \kappa(0) - \rho^2 / (\kappa(0) + \sigma_{\omega}^2 / |\mathbb{I}_i(t_k,\rho)|)$. 
    Therefore, it is straightforward to see that the error bound decreases monotonically with respect to both $|\mathbb{I}_i(t_k,\rho)|$ and $\rho$, while increases with the noise variance $\sigma_{\omega}^2 $. 
    Intuitively, increasing the selection variable $\rho$ requests more related data points, however, it may not enlarge the number of $|\mathbb{I}_i(t_k,\rho)|$.
    Therefore, with the fixed selection variable $\rho$, the bigger set $|\mathbb{I}_i(t_k,\rho)|$ signifies higher data density. 
    Meanwhile, the small noise variance $\sigma_{\omega}^2$ indicates the high quality of training data, leading to more reliable predictions. 
\end{remark}

Based on \cref{lemma_prediction_error}, we are able to derive the prediction error bound of the individual GP model using EIGP and to evaluate the overall cooperative learning performance within the MAS.

\begin{theorem} \label{theorem_aggregated_prediction_error_bound}
Let all assumptions in \cref{lemma_prediction_error} hold and use the aggregation method with the form \eqref{eq_EIGP_framework} using the aggregation weights either \eqref{eqn_aggregation_weights_gEIGP} or \eqref{eqn_aggregation_weights_aEIGP}.
Then, the error of the aggregated prediction $\tilde{\bm{\mu}}_i(\bm{x}(t_k)) = [\tilde{\mu}_i^1(\bm{x}(t_k)), \cdots, \tilde{\mu}_i^d(\bm{x}(t_k))]^\top$ is bounded as
\begin{align} \label{eqn_aggregated_prediction_error_bound}
    \| \tilde{\bm{\mu}}_i(\bm{x}(t_k)) - \bm{f}(\bm{x}(t_k)) \| \le \tilde{\eta}_i(\bm{x}(t_k)),
\end{align}
with a probability of at least $1 - n \delta_x$, where  $\tilde{\eta}_i(\bm{x}(t_k)) = \| [\tilde{\eta}_i^1(\bm{x}(t_k)), \cdots, \tilde{\eta}_i^d(\bm{x}(t_k))]\|$. 
Moreover, define $\hat{\eta}_i^j(\bm{x}(t_k))= \sum_{s \in \mathcal{S}_i(\boldsymbol{x}(t_k))} w_{is}^{j}(\boldsymbol{x}(t_k)) \eta_i^j(\bm{x}(t_k))$  with
\begin{align}
    \tilde{\eta}_i(\bm{x}(t_k)) &=   \varepsilon_i^{-1}(\boldsymbol{x}(t_k)) \|\tilde{\boldsymbol{\mu}}_{i}(\bm{x}(t_k) ) \|  + \eta_i(\bm{x}(t_k)),
\end{align}
for $j = 1,\dots, d$. Then, the prediction error of all aggregated predictions in MAS is bounded as
\begin{align} \label{eqn_aggregated_prediction_error_bound_all}
    \| \hat{\bm{f}}(\bar{\bm{x}}(t_k)) - &\bm{f}(\bar{\bm{x}}(t_k)) \| \le \hat{\eta}(\bar{\bm{x}}(t_k)) \\
    &= \| [\hat{\eta}_1(\bm{x}_1(t_k)), \cdots, \hat{\eta}_n(\bm{x}_n(t_k))]\|,\nonumber
\end{align}
where $\bm{f}(\bar{\bm{x}}(t_k)) = [\bm{f}(\bm{x}_1(t_k)), \cdots, \bm{f}(\bm{x}_n(t_k))]$, $\tilde{\bm{\mu}}(\bar{\bm{x}}(t_k)) = [\tilde{\bm{\mu}}_1(\bm{x}_1(t_k)), \cdots, \tilde{\bm{\mu}}_n(\bm{x}_n(t_k))]$ and $\bar{\bm{x}} = [\bm{x}_1,\dots,\boldsymbol{x}_n]$, with probability of at least $1 - n^2 \delta_x$.
\end{theorem}
 \begin{IEEEproof}
    With triangular inequality and the result in Lemma \ref{lemma_prediction_loss}, we have
	\begin{align} \label{eqn_prediction_error_derive}
		\| &\tilde{\bm{\mu}}_i(\bm{x}(t_k)) - \bm{f}(\bm{x}(t_k)) \| \nonumber \\
  &\le \| \tilde{\bm{\mu}}_i(\bm{x}(t_k)) - \bm{\mu}_{i}(\bm{x}(t_k)) \| + \| \bm{\mu}_{i}(\bm{x}(t_k)) - \bm{f}(\bm{x}(t_k)) \| \nonumber\\
		&\le \varepsilon_i^{-1}(\bm{x}(t_k)) \| \tilde{\bm{\mu}}_{i}(\bm{x}(t_k)) \| + \eta_i(\bm{x}(t_k)) = \tilde{\eta}_i(\bm{x}(t_k))
	\end{align}
	with a probability of at least $1 - \delta_x$ considering the union bound. From the expression of $w_{is}^j$ in \eqref{eqn_aggregation_weights_gEIGP} or \eqref{eqn_aggregation_weights_aEIGP} , it is obvious $\sum_{s \in \mathbb{S}_i(\boldsymbol{x})} w_{is}^{j}(\boldsymbol{x}) = 1$, such that
    \begin{align}
        \|& \hat{\bm{f}}_i(\bm{x}(t_k)) - \bm{f}(\bm{x}(t_k))\| \nonumber \\
        &\le \sum_{s \in \mathbb{S}_i(\boldsymbol{x}(t_k))} w_{is}(\boldsymbol{x}(t_k)) \| \tilde{\bm{\mu}}_i(\bm{x}(t_k)) - \bm{f}(\bm{x}(t_k)) \| \nonumber \\
        &\le \sum_{s \in \mathbb{S}_i(\boldsymbol{x}(t_k))} w_{is}(\boldsymbol{x}(t_k)) \tilde{\eta}_i(\bm{x}(t_k)) = \hat{\eta}_i(\bm{x}(t_k)),
    \end{align}
   holds for any dimension $j = 1, \cdots, m$ with the probability of at least $1 - n \delta_x$ using the union bound, where the second inequality is obtained using the result in \cref{lemma_prediction_error}.
Use the triangle inequality considering $\| \tilde{\bm{\mu}}_i(\bm{x}(t_k)) - \bm{f}(\bm{x}(t_k)) \|$ is the concatenation of $\|\hat{\bm{f}}_i(\bm{x}(t_k)) - \bm{f}(\bm{x}(t_k))\|$, then the result in \eqref{eqn_aggregated_prediction_error_bound} is straightforwardly obtained. And it can be extended to \eqref{eqn_aggregated_prediction_error_bound_all} by considering
    \begin{align}
        \| &\hat{\bm{f}}(\bm{x}(t_k)) - \bm{1}_n \otimes \bm{f}(\bm{x}(t_k)) \|^2 \nonumber \\
        &= \sum_{i=1}^n \|\hat{\bm{f}}_i(\bm{x}(t_k)) - \bm{f}(\bm{x}(t_k))\|^2 \\
        &\le \sum_{i=1}^n \hat{\eta}_i^2(\bm{x}(t_k)) = \| [\hat{\eta}_1(\bm{x}(t_k)), \cdots, \hat{\eta}_n(\bm{x}(t_k))]^\top \|^2, \nonumber
    \end{align}
    with the probability of at least $1-n^2\delta_x$ due to the union bound.
\end{IEEEproof}

\section{Numerical Demonstration}\label{sec_sim}
We present a series of simulations\footnote{Codes and data sets all are available in the supplementary material.} across offline learning and online learning to empirically demonstrate the effectiveness of the proposed methodology in comparison to existing state-of-the-art distributed GP approaches in the following subsections. 

\subsection{Toy Example}\label{subsec_toyExample}
\begin{figure}[htp]
\centering 

\subfloat{%
  \label{fig_toyData}\includegraphics[clip,width=0.9\columnwidth]{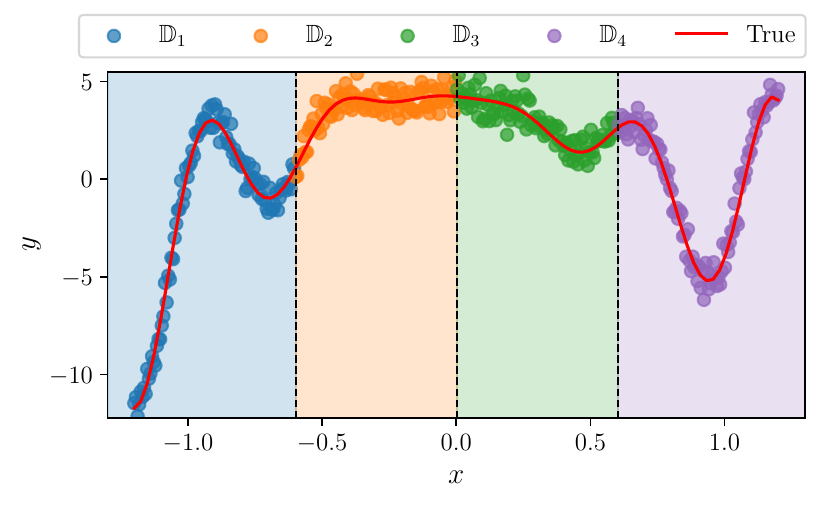}%
}
\vspace{5pt}
\subfloat{%
  \label{fig_agentQuantity_toy}\includegraphics[clip,width=0.9\columnwidth]{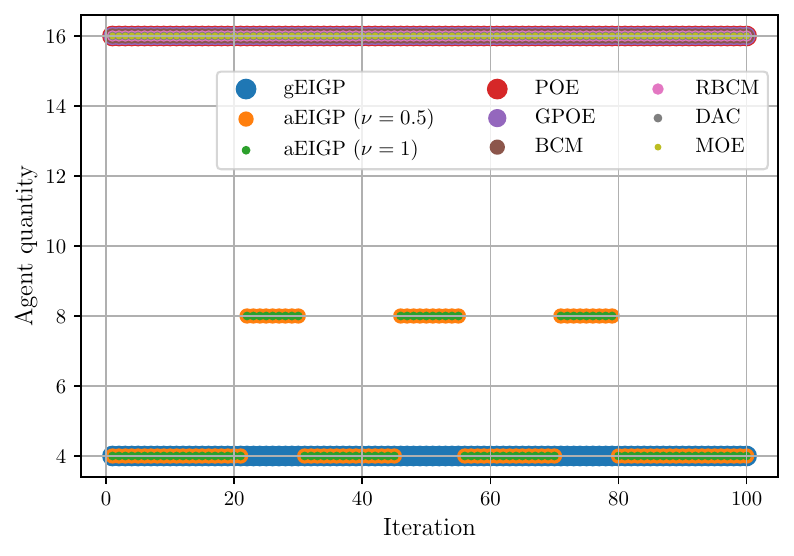}%
}

\caption{Training data sets, true function, and agent quantity. Training data sets and true value of the function (above). The overall quantity of the agents included within the MAS for each iteration (below).}

\end{figure}

To elucidate the comprehensive understanding of the proposed EIGP, we first adopt an offline regression task to showcase the prediction performance in a MAS with 4 agents $(n=4)$ by following the same example in \cite{liu2018generalized}. 
Each agent is characterized by a distinct training dataset (see \cref{fig_toyData}) and communicates through a fully connected network.  The function is defined as 
$$f(x) \!=\! 5x^2 \sin(12x) + (x^3 - 0.5) \sin(3x - 0.5) + 4 \cos(2x) + \epsilon_0,$$ 
where the noise term $\epsilon_0\sim \mathcal{N}(0, 0.25)$. We employ a dataset comprising 400 training points uniformly distributed in the interval $[-1.2, 1.2]$, partitioned into distinct subsets for each agent. Specifically, the $i$-th agent possesses the data set $\mathbb{D}_i$, $i=1,2,3,4$ (refer to \cref{fig_toyData}). 
To evaluate the regression performance, we assess 100 query points across the interval. 
The overall quantity of agents involved within the MAS for each iteration is depicted in \cref{fig_agentQuantity_toy}. 

 \begin{figure}[h]
    \centering
    \includegraphics[scale=0.49]{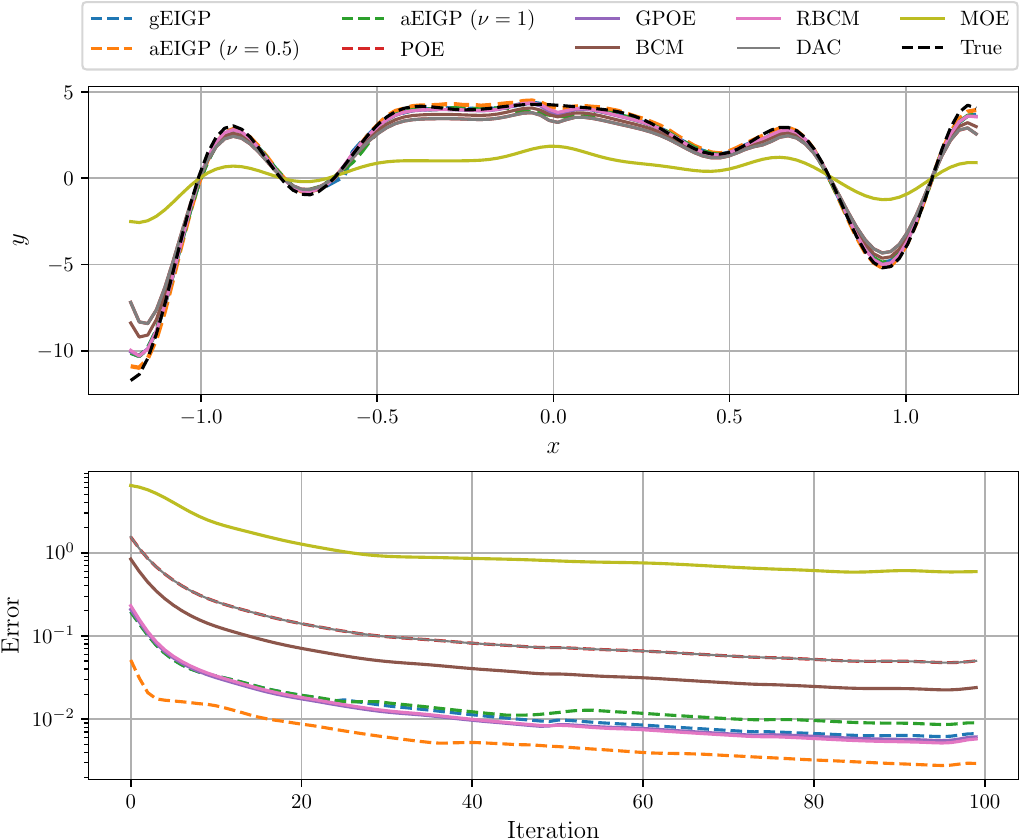}
    \vspace{-0.3cm}
    \caption{Predictions of different methods (above) and the absolute errors between predictions and true values. (bottom). }
    \vspace{-0.3cm}
    \label{fig_toyStates}
\end{figure}

Notably, gEIGP involves only four agents throughout the entire simulation indicating each agent only selects one GP model to make predictions. This is attributed to the selection process wherein each agent opts for the highest quality model for prediction. Meanwhile, aEIGP involves two agents near the division boundary (black dash line in \cref{fig_toyData}) and only requires one agent for predictions within the division intervals. This underscores a wise selection of agents when necessary. Conversely, other methodologies that engage all available agents for predictions yield diminished prediction accuracy and prolonged computational times, underscoring the inadequacy of indiscriminate agent involvement in prediction tasks.

\begin{figure}[ht]
    \centering
    \includegraphics[scale=0.35]{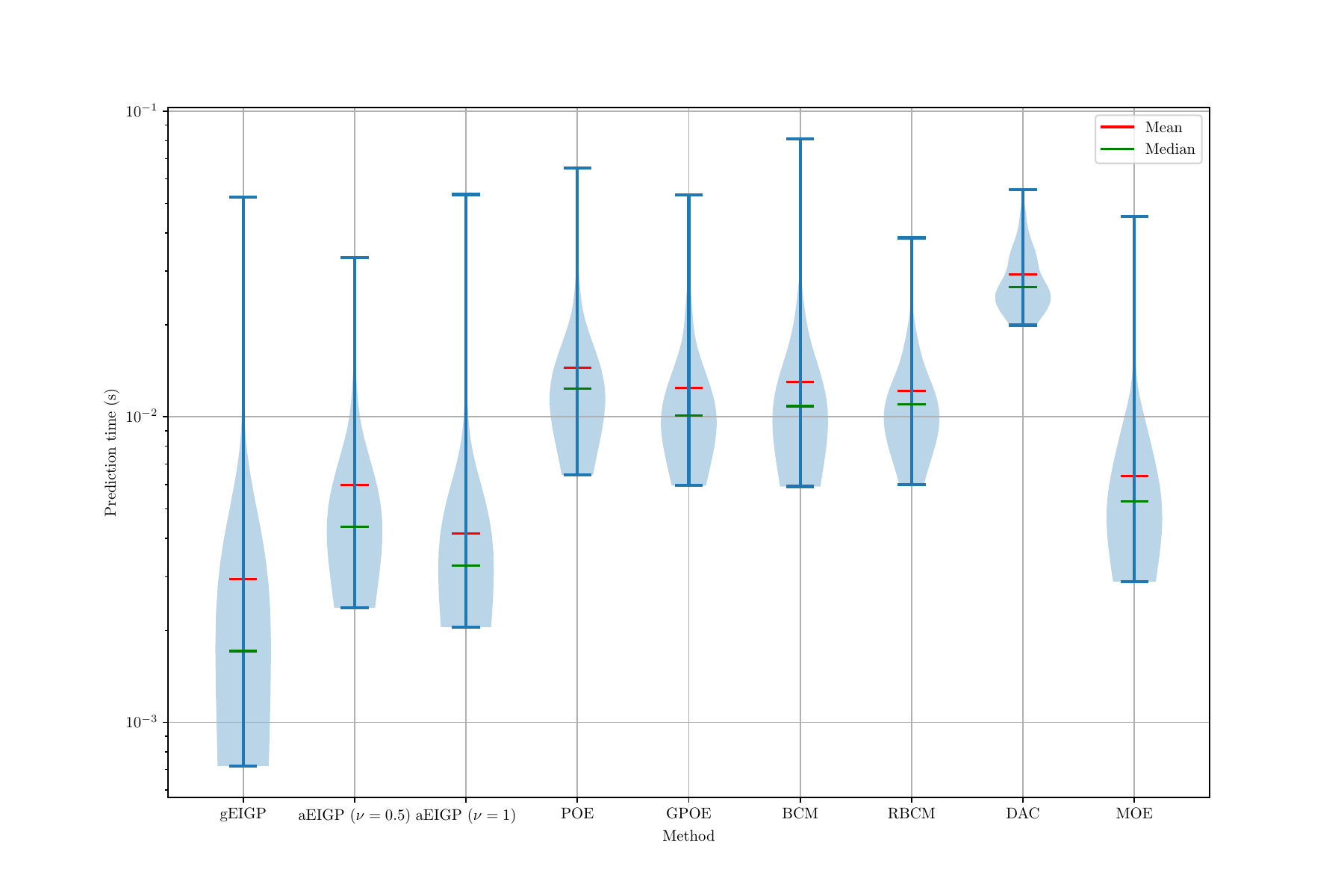}
    \vspace{-0.3cm}
    \caption{Prediction time for each iteration.}
    \vspace{-0.3cm}
    \label{fig_predictionTime_toy}
\end{figure}

For each prediction instance within the MAS, a violin plot is generated, as illustrated in \cref{fig_predictionTime_toy}. This visualization provides insight into the distribution of prediction times across 100 iterations. The plots demonstrate that all variants of EIGP exhibit shorter computational time in both mean and median values. \cref{fig_toyStates} demonstrates the superior performance of aEIGP ($\nu=0.5$) when compared to alternative approaches. Notably, gEIGP exhibits comparable performance to GPOE and RBCM. Additionally, the computational time efficiency of EIGP methods is underscored in \cref{table_predictionTime}, where all EIGP variants require less time for prediction calculations. 
\cref{fig_agentQuantity_toy} illustrates the total number of agents engaged within the MAS for each prediction iteration. There are only four agents involved, signifying that each agent requires just one agent for prediction when utilizing gEIGP. Meanwhile, aEIGP necessitates two agents during the division boundary and only requires one agent for predictions within the division, highlighting the rational selection of agents when needed. Conversely, other methods involving the use of all available agents for predictions result in inferior prediction accuracy and longer computational times, emphasizing the inadequacy of indiscriminate agent involvement in prediction.

\begin{table*}
\caption{The mean prediction time (ms) of the MAS for each iteration. \textbf{Bold} numbers signify the first smallest values and \textit{italics} the second. }
    \centering
    \begin{tabular}{lrrrrrrrrrrr}
        \toprule
        Scenario  & gEIGP & aEIGP $(\nu=\frac{1}{2})$& aEIGP $(\nu={1})$& POE& GPOE& BCM& RBCM& DAC& MOE \\
        \midrule
        Toy ($n$=4)     & \textbf{2.94}  & 5.97& \textit{4.15}& 14.48& 12.45& 13.01& 12.15& 29.18& 6.41        \\

        KIN40K ($n$=8) & \textbf{11.47}          & 24.09& \textit{18.52}& 41.64& 40.14  & 43.92& 41.95& 290.70& 21.28      \\
        KIN40K ($n$=16) & \textbf{35.68}          & 60.63& \textit{47.34}& 155.12& 154.12  & 152.04& 155.62& 789.80& 78.52      \\
        POL ($n$=8) &  \textbf{6.98} & 15.80 & \textit{13.33}& 45.13& 42.04  & 41.75& 42.98& 264.51& 22.00      \\
        POL ($n$=16) & \textbf{23.90}          & 41.52& \textit{35.13}& 159.40& 155.82  & 153.10& 157.47& 871.43& 80.31      \\
        \bottomrule
    \end{tabular}
    
    \label{table_predictionTime}
\end{table*}


\subsection{Benchmarks}\label{subsec_regressionBenchmarks}
For evaluating the learning performance of EIGP in the online scenario, we selected two real-world datasets (KIN40K and POL), each comprising 10$^4$ data points. 
In both scenarios with MAS configurations of $8$ and $16$ agents communicate via a fully connected network and receive the streaming data sequentially with $100$ data threshold. 
Data points were sequentially added, with each agent accumulating data until reaching its threshold, after which the subsequent agent would commence data acquisition. Once the final agent reached its limit, the process resumed from the initial agent, ensuring a cyclic data augmentation process.
This sequential data augmentation strategy ensures a controlled and realistic simulation environment for assessing online learning performance within the MAS.

\begin{figure}[ht]
    \centering
    \includegraphics[scale=0.49]{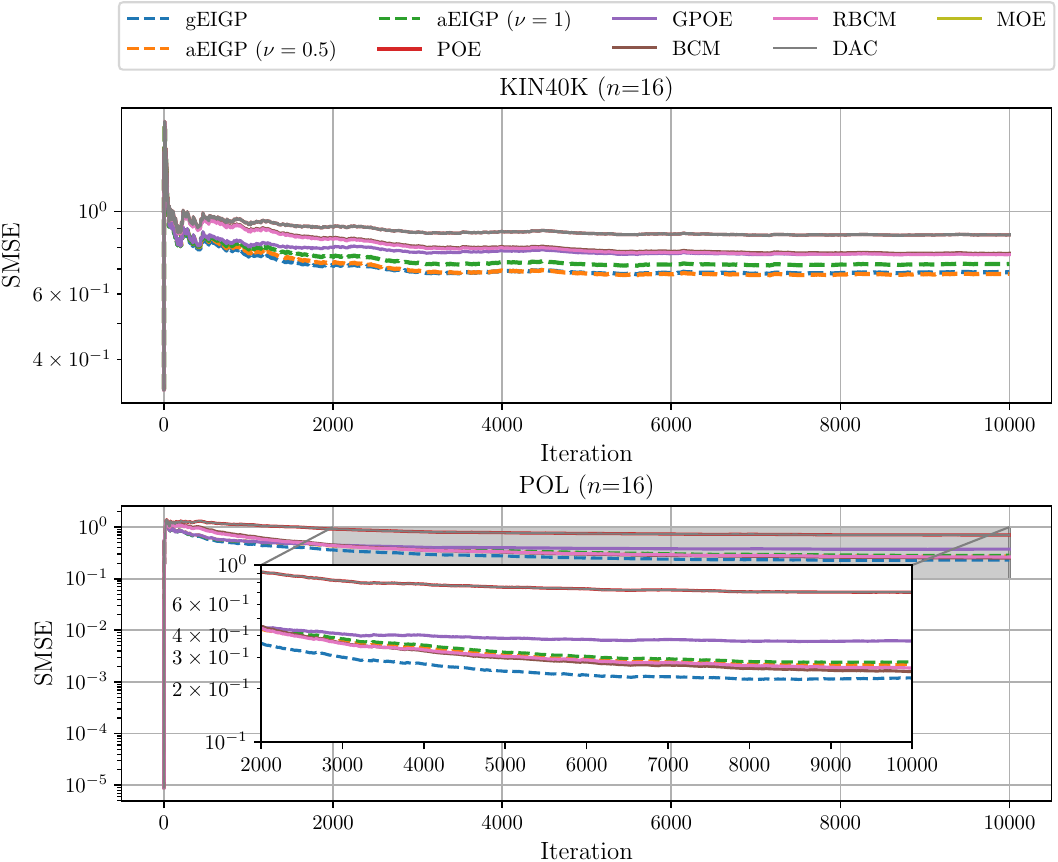}
    \vspace{-0.3cm}
    \caption{The plots of SMSE of each iteration with KIN40K (above) and POL (bottom) data sets.}
    \vspace{-0.3cm}
    \label{fig_benchmerks_16agents}
\end{figure}

\begin{figure}[h]
    \centering
    \includegraphics[width=1\linewidth]{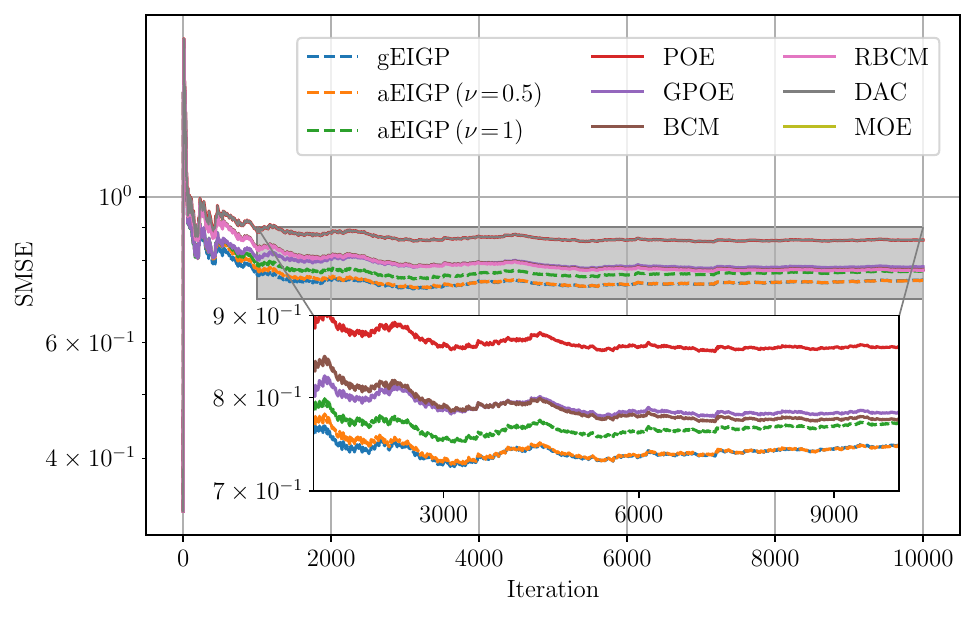}
    \caption{The plots of SMSE of each iteration with KIN40K data sets in the 8-agent MAS.}
    \label{fig_benchmerks_8agents_KIN40K}
\end{figure}

\begin{figure}[h]
    \centering
    \includegraphics[width=1\linewidth]{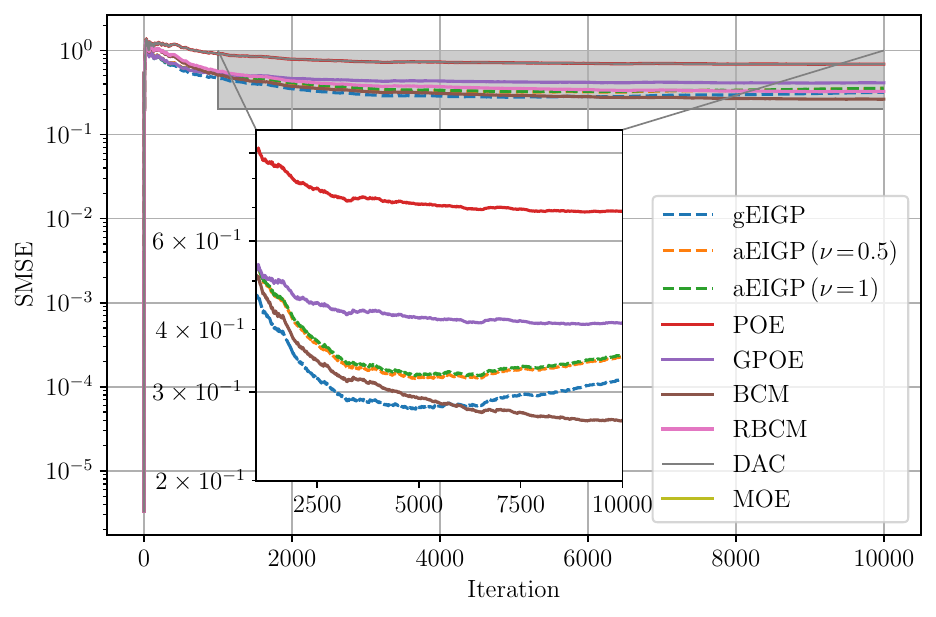}
    \caption{The plots of SMSE of each iteration with POL data sets in the 8-agent MAS.}
    \label{fig_benchmerks_8agents_POL}
\end{figure}

In the MAS with $16$ agent scenario\footnote{For simulation configuration and additional results, refer to \cref{subsec_Benchmarks}.}, \cref{fig_benchmerks_16agents} shows all EIGP variants outperform alternative methods in the overall standardized mean squared error (SMSE) of the MAS in the KIN40K data set. Meanwhile, gEIGP demonstrates superior performance in the POL data set, and aEIGP exhibits a similar performance with BCM and RBCM. 
 For $8$ agent scenario, all EIGP variants have small SMSE in the KIN40K dataset during the overall time in \cref{fig_benchmerks_8agents_KIN40K}. However, in the POL dataset, \cref{fig_benchmerks_8agents_POL}gEIGP demonstrates superior performance until the 6000th iteration, after which BCM exhibits a comparable yet slightly improved performance, and aEIGP exhibits a similar performance with RBCM. 
\cref{table_predictionTime} shows that, across all scenarios, gEIGP exhibits the shortest mean prediction time due to the fact that gEIGP only utilizes the optimal agent to predict, with aEIGP ($\mu=1$) closely following suit.

\begin{figure*}[h]
    \centering
    \subfloat{{\includegraphics[width=1\columnwidth]{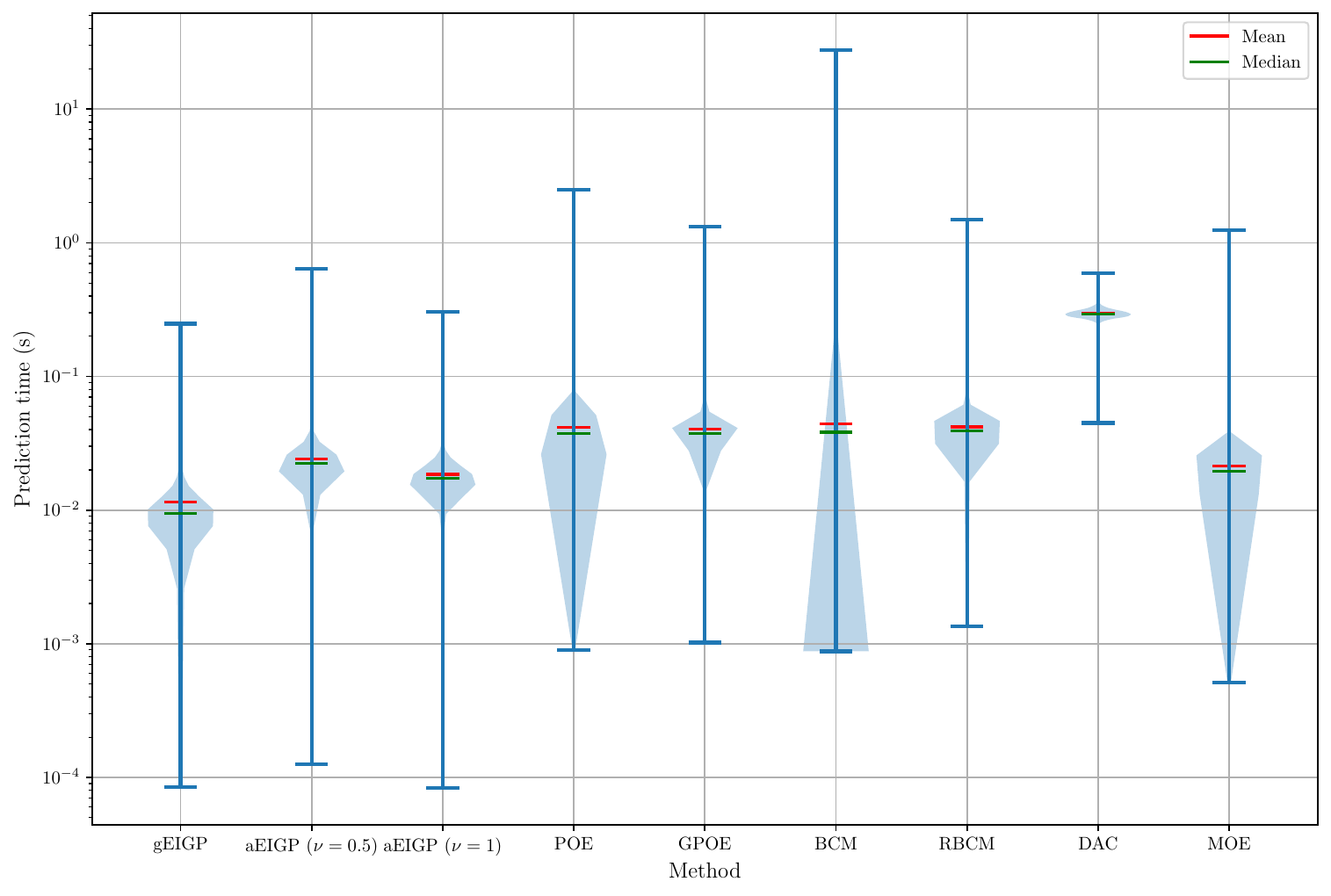} }}%
\hspace{0.01cm}
    \subfloat{{\includegraphics[width=1\columnwidth]{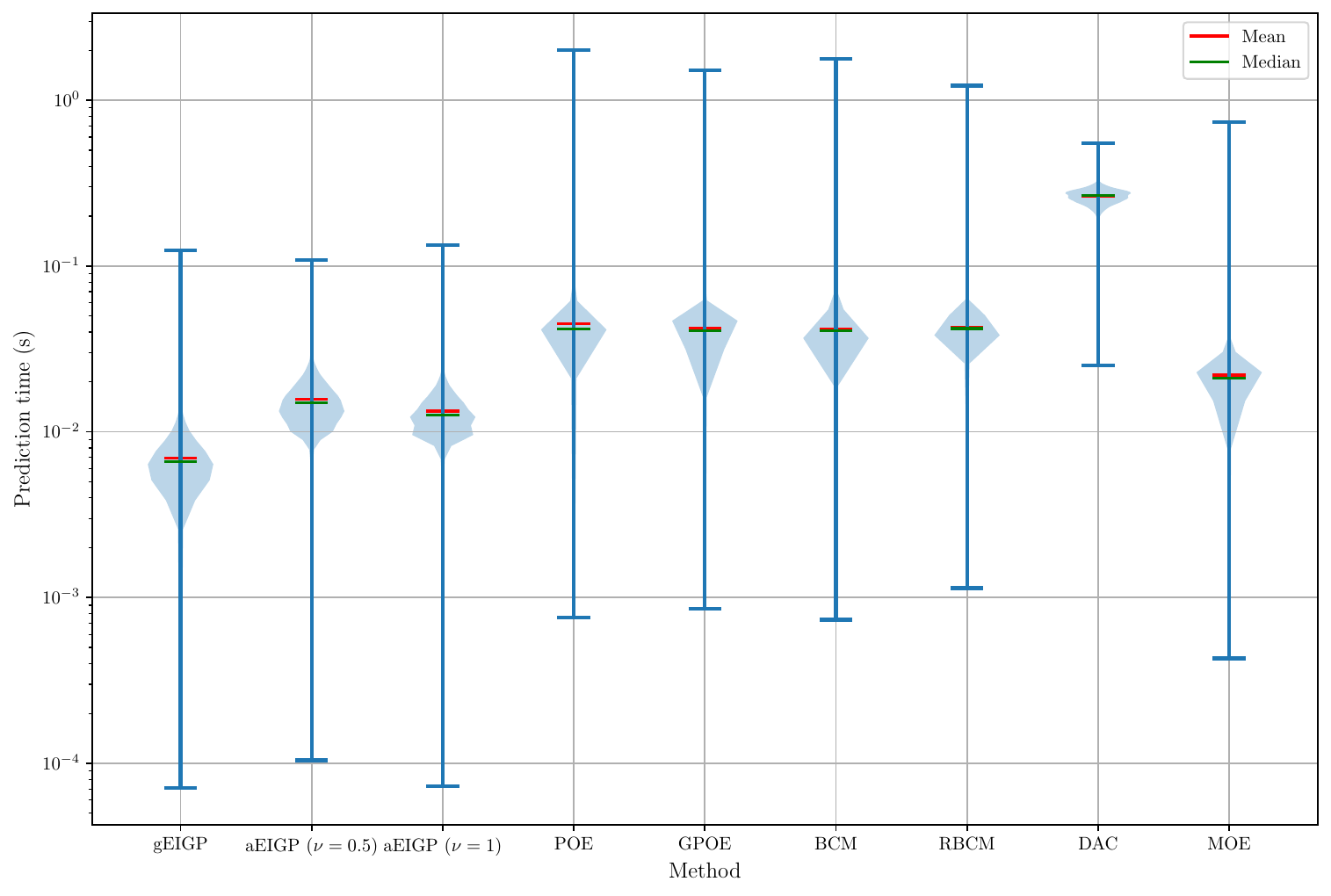} }}%
    \caption{Prediction time for each iteration of the MAS with 8 agents in KIN40K (left) and POL
(right) data sets.}%
    \label{fig_predictTime_8}%
\end{figure*}

\begin{figure*}[h!]
    \centering
    \subfloat{{\includegraphics[width=1\columnwidth]{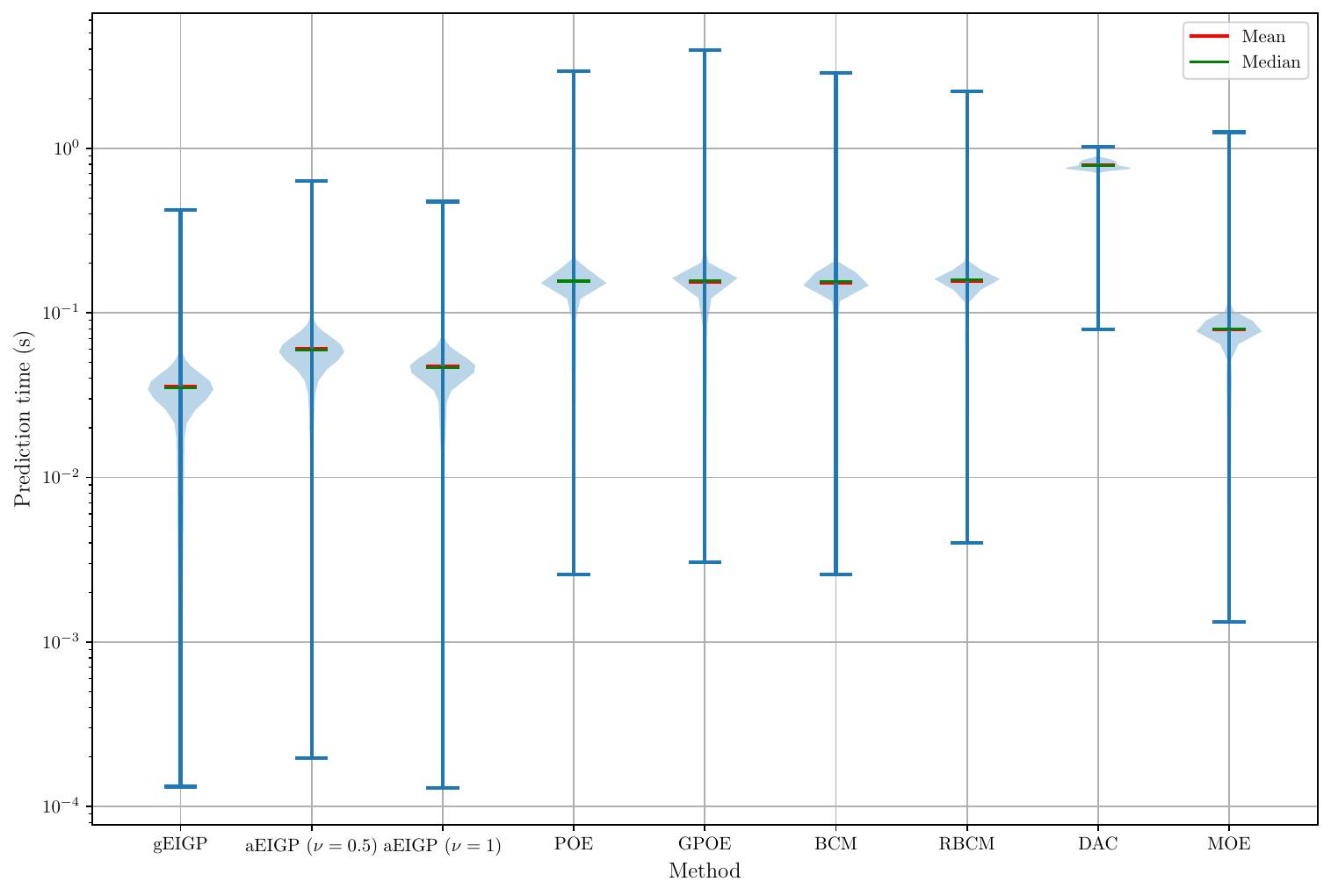} }}%
\hspace{0.01cm}
    \subfloat{{\includegraphics[width=1\columnwidth]{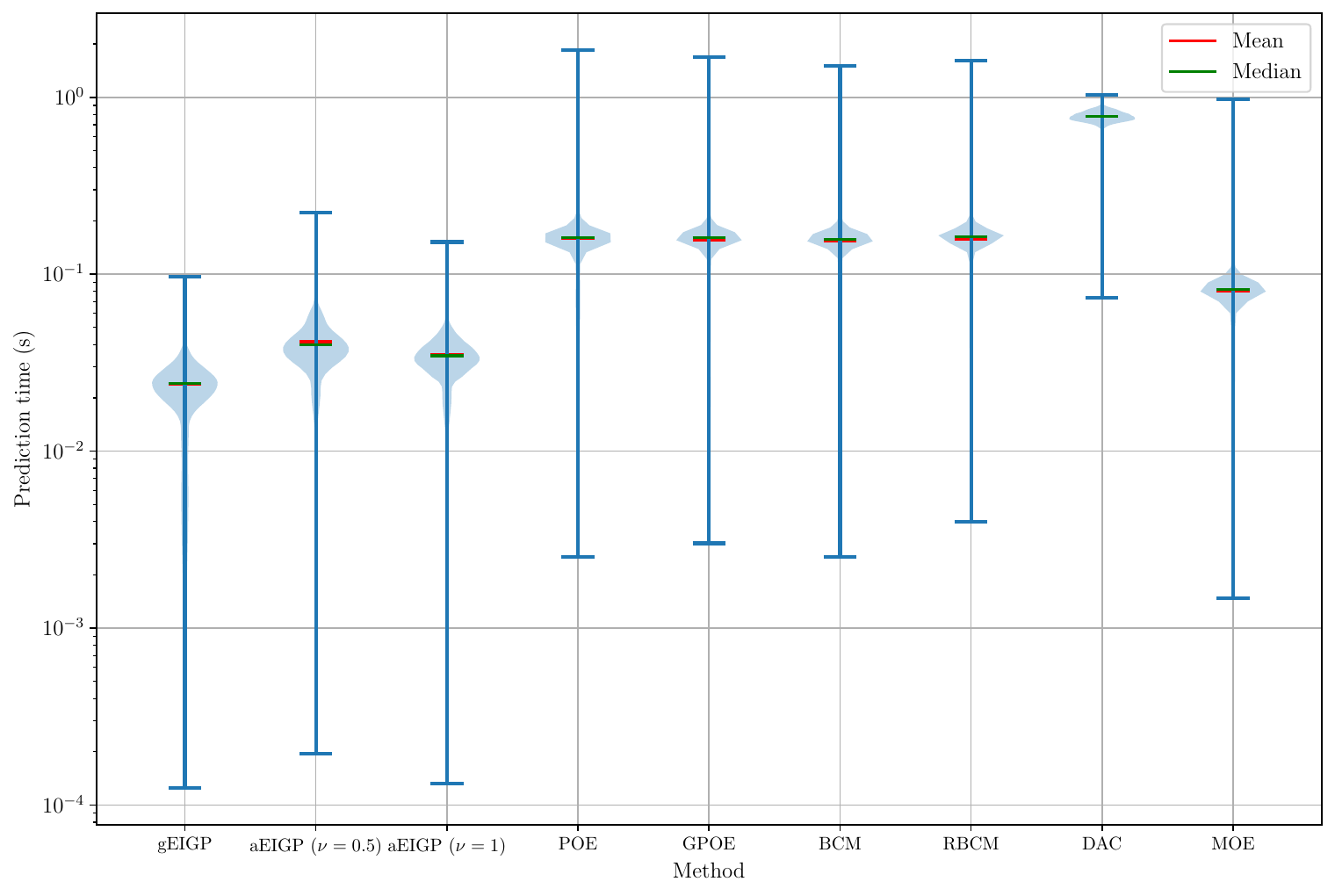} }}%
    \caption{Prediction time for each iteration of the MAS with 16 agents in KIN40K (left) and POL
(right) data sets.}%
    \label{fig_predictTime_16}%
\end{figure*}

In order to demonstrate the distribution of prediction times during the 10$^4$ interactions for both the scenarios of $8$ agents and $16$ agents in two data sets. We present the violin plots in \cref{fig_predictTime_8} and \ref{fig_predictTime_16}. It is evident that gEIGP consistently demonstrates the shortest mean and median prediction times. Following closely is aEIGP ($\mu=1$), which consistently ranks as the second shortest in terms of prediction time. Similarly to toy example, it is evident that gEIGP consistently demonstrates the shortest mean and median prediction times across all examples. Following closely is aEIGP ($\mu=1$), which consistently ranks as the second shortest in terms of prediction time.

Figures \ref{fig_kin40k_agentQuantity_8} to \ref{fig_pol_agentQuantity_16} demonstrate the involvement of agents in each prediction within the MAS. Significantly, it is discernible that the proposed EIGP methods obviate the requirement for the participation of all agents in each prediction. In every iteration, gEIGP consistently engages only one agent for making predictions, thereby reflecting its efficiency in terms of reduced computational time.
Moreover, the aEIGP methods selectively activate agents solely when their contribution is necessary. This activation strategy aligns with the overarching theme of prioritizing model quality over the sheer quantity of agents in the cooperative learning.

\section{Conclusion}\label{sec_conclusion}
This paper presents a novel distributed learning framework, EIGP, which facilitates intelligent agent selection for cooperative learning. Moreover, the proposed EIGP algorithms optimize computational efficiency by selecting fewer agents in joint predictions and simultaneously improve prediction accuracy by using high-quality agents. 
The analysis of uniform error bounds across predictions within the MAS provides the guarantees of prediction performance, which plays a particularly important role in safety-critical applications. 
The simulations demonstrate the effectiveness of EIGP and its adaptability to real-time online learning tasks.

While our current data management strategy prioritizes computational efficiency, one limitation of the EIGP is that the model requires updates after new data is added, which can result in delayed predictions in real-time applications. 
To mitigate this, future work will integrate with state-of-the-art asynchronous distributed GP frameworks~\cite{yang_AAAI2025_asynchronous}, to resolve the asynchronous prediction problem.
Additionally, another limitation of the EIGP is its reliance on a predefined parameter $\theta$ in \cref{eq_phi_aEIGP}, which controls the confidence interval range in the aEIGP algorithm. 
This static threshold may not always adapt effectively. 
Therefore, future research will investigate more adaptive thresholding mechanisms to balance the number of selective neighbors and computational time.

\appendices
\section{Generalization of aEIGP}\label{sec_aEIGP_variants}
As established in prior literature, the posterior variance in \cref{eqn_GP_variance} of GP regression, functions as a reliable indicator of prediction uncertainties~\cite{hintonTrainingProductsExperts2002,ng2014hierarchical,yangDistributedLearningConsensus2021,ledererCooperativeControlUncertain2023}. Consequently, it exists an opportunity to incorporate this metric with the proposed error-informed aggregation weight $\tilde{w}_{is}^j$ in \cref{eq_tildeW}. This strategic fusion allows for a more comprehensive and informed approach to the determination of aggregation weights, contributing to a robust aggregation for enhancing the overall predictive accuracy and reliability of the model in the overall predictive modeling framework, which can be designed by
\begin{align}\label{eq_aEIGP_weight_generaliztion}
    {w}_{is}^j(\boldsymbol{x})  = \Psi_{is}\big(\cup_{s\in\mathcal{S}_i} \Phi (\tilde{w}_{is}^j(\boldsymbol{x}), \varpi_{is}( \boldsymbol{\sigma}_{i}^j(\boldsymbol{x})) ,\nu)\big),
\end{align}
where $\Psi_{is}$ is the $s$-th element of the proportional normalization function, $\Phi$ represents a trade-off function that determines the contribution of the variables $\tilde{w}_{is}^j$ and $\varpi_{is}$ according to the weighting factor $\nu\in [0,1]$, and posterior variance set $\boldsymbol{\sigma}_{i}^j(\boldsymbol{x}) = \{{\sigma}_{is}^j(\boldsymbol{x})\}_{s\in\mathcal{S}_i}$. Specifically, consider a set $\{{v}_{is}\}_{s\in\mathcal{S}_i}$, the $\Psi_{is}(\cdot)$ is defined by
\begin{align}\label{eq_proportional}
    \Psi_{is}(\cup_{s \in \mathcal{S}_i} v_{is}) = \frac{v_{is}}{\sum_{l \in \mathcal{S}_i} v_{il}}.
\end{align}
This normalization function operates by proportionally normalizing the value $v_{is}$ within the context of a set $\mathcal{S}_i$. 
Each element $v_{is}$ is divided by the sum of all elements in the set $\mathcal{S}_i$. 
The outcome of this normalization process is a set of values that maintain a proportional relationship to one another, thereby facilitating a balanced comparison and interpretation of their magnitudes within the specified set. 
It holds significance in ensuring that the results of the normalization process collectively sum to 1. 

The function $\Phi(a,b,\nu)$ represents a trade-off mechanism between two variables, i.e., $a$ and $b$, incorporating a tuning factor governed by $\nu$. In essence, the parameter $\nu$ operates as a tuning variable, affording the user the flexibility to tailor the trade-off between variables. We list some common functions $\Phi(a,b,\nu)$ as follows.
\begin{enumerate}[label=(\alph*)]
\item Linear trade-off function: 
    \begin{align}\label{eq_tradeoff_linear}
    \Phi(a, b, \nu)_{\text{Lin}} = \nu a + (1-\nu) b
    \end{align}
\item Power trade-off function: 
    \begin{align}
    \Phi(a, b, \nu)_{\text{Pow}} =  a^{\nu} b^{(1-\nu)}
    \end{align}
\item Exponential trade-off function:
    \begin{align}
    \Phi(a, b, \nu)_{\text{Ex}} =  \exp (\nu a) + \exp ((1-\nu) b)
    \end{align}
\item Logarithmic trade-off function:
    \begin{align}
        \Phi(a, b, \nu)_{\text{Log}} = \nu \log(a) + (1-\nu) \log(b) 
    \end{align}
\end{enumerate}

The $\varpi_{is}$ depends on the variances of GP regression. We summarize the variance-based aggregation weight functions as follows
\begin{enumerate}[label=(\alph*)]
    \item POE family~\cite{hintonTrainingProductsExperts2002,ng2014hierarchical,caoGeneralizedProductExperts2015}:
    \begin{align}\label{eq_aggregation_poe}
        \varpi_{is}( \boldsymbol{\sigma}_{i}^j(\boldsymbol{x}))  = \frac{\vartheta_{is}^j\sigma_{is}^j(\boldsymbol{x})}{\sum_{l \in \mathcal{S}_i} \vartheta_{il}^j\sigma_{il}^j(\boldsymbol{x})},
    \end{align}
    where $\gamma_{is}\in\mathbb{R}_{+}$ is a predefined parameter. \cite{caoGeneralizedProductExperts2015} suggest setting $\vartheta_{il}^j$ to the difference in the differential entropy between the prior and the posterior to determine the importance.
    \item BCM family~\cite{trespBayesianCommitteeMachine2000,deisenrothDistributedGaussianProcesses2015,liu2018generalized}:
    \begin{align}\label{eq_aggregation_bcm}
         \varpi_{is}( \boldsymbol{\sigma}_{i}^j(\boldsymbol{x})) =  \frac{\vartheta_{is}^j\sigma_{is}^j(\boldsymbol{x})}{\sum_{l \in \mathcal{S}_i} \sigma_{il}^j(\boldsymbol{x}) + (1-\sum_{l\in\mathcal{S}_i} \vartheta_{il}^j ) \sigma_{\text{prior}}},
    \end{align}
    where $\sigma_{\text{prior}}$ is the prior variance of GPs. Moreover, \cite{liu2018generalized} suggest consider the noise of the training dataset, which leads the second term in the denominator in \cref{eq_aggregation_bcm} to be $(1-\sum_{l\in\mathcal{S}_i} \vartheta_{il}^j )(\sigma_{\text{prior}} + \sigma_{\omega})$.
\end{enumerate}

Up to this point, we have formulated the general framework of the aEIGP algorithm. Given the numerous combinations available in the co-design of trade-off functions $\Phi$ and variance-based aggregation functions $\varpi$. Specifically, we opt for a power trade-off function due to the empirical results analysis. In addition, the variance-based aggregation function is designed under the consideration of the prior variance of GPs, which is motivated by the BCM family approaches. Consequently, by defining $\vartheta_{il}^j(\boldsymbol{x}) = \log((\kappa(0) +{\sigma_{\omega}^{j}}^2)/{\sigma_{is}^{j}}^{2}(\boldsymbol{x}))$, we unveil the aggregation weights of the aEIGP algorithm as follows
\begin{align}
    &w_{is}^{j}(\boldsymbol{x}) \!=\!  \frac{\big(\tilde{w}_{is}(\boldsymbol{x})\big)^{\nu} \Big(\log\big(\frac{\kappa(0) +{\sigma_{\omega}^{j}}^2}{{\sigma_{is}^{j}}^{2}(\boldsymbol{x})}\big){\sigma_s^{j}}^{-2}(\boldsymbol{x})\tilde{\sigma}{{}_i^{j}}^2(\boldsymbol{x}) \Big)^{(1-\nu)} }  {\sum_{l\in{\mathcal{S}}_i(\boldsymbol{x})}  \big(\tilde{w}_{il}(\boldsymbol{x})\big)^{\nu}\! \Big( \log\big(\frac{\kappa(0) +{\sigma_{\omega}^{j}}^2}{{\sigma_{is}^{j}}^{2}(\boldsymbol{x})}\big) \frac{\tilde{\sigma}{{}_i^{j}}^2\!(\boldsymbol{x})}{{\sigma_l^{j}}^{2}(\boldsymbol{x})} \Big)^{(1-\nu) } },\\
&\tilde{\sigma}{{}_i^{j}}^{-2}(\boldsymbol{x})  \!=\! \!\sum_{l\in\mathcal{S}_i(\boldsymbol{x})} \tilde{w}_{il}(\boldsymbol{x}) \log((\kappa(0) +{\sigma_{\omega}^{j}}^2)/{\sigma_{il}^{j}}^{2}(\boldsymbol{x})){\sigma_l^j}^{-2}(\boldsymbol{x}) \nonumber\\
&+\! \Big(1- \!\sum_{l\in\mathcal{S}_i(\boldsymbol{x})} (\tilde{w}_{is}(\boldsymbol{x}))^{\nu} ( \vartheta_{is}^j(\boldsymbol{x}))^{(1-\nu)}\Big) / (\kappa(0)  + {\sigma_{\omega}^{j}}^2 ).\!\!
\end{align}
To make an example, if we chose the linear trade-off function \eqref{eq_tradeoff_linear} and POE family \eqref{eq_aggregation_poe} with $\gamma_{is}=1$, the aEIGP-LinPOE is formulated as
\begin{align}
    &w_{is}^{j}(\boldsymbol{x}) \!=\!  \frac{{\nu}\tilde{w}_{is}(\boldsymbol{x})+  (1-\nu){\sigma_s^{j}}^{-2}(\boldsymbol{x})\tilde{\sigma}{{}_i^{j}}^2(\boldsymbol{x})}  {\sum_{l\in{\mathcal{S}}_i(\boldsymbol{x})}  {\nu}\tilde{w}_{il}(\boldsymbol{x}) + (1-\nu) {\sigma_l^{j}}^{-2}\!(\boldsymbol{x})\tilde{\sigma}{{}_i^{j}}^2(\boldsymbol{x}) },\\ &\tilde{\sigma}{{}_i^{j}}^{-2}(\boldsymbol{x}) = \sum_{l\in\mathcal{S}_i(\boldsymbol{x})} \tilde{w}_{il}(\boldsymbol{x}) {\sigma_l^j}^{-2}\!(\boldsymbol{x}),
\end{align}
It is imperative to recognize that the incorporation of posterior-based aggregation weights in \cref{eq_aggregation_poe} and \cref{eq_aggregation_bcm}, introduces additional computational demands for collaborating agents. Moreover, this is accompanied by an increased volume of information exchange to the posterior variances. However, these adjustments yield a source for predictive information from the collaborating agents.

It is noteworthy that the aEIGP framework offers an avenue for achieving adaptability, specifically when setting $\nu=1$. 
This particular choice implies that there is no need for the inclusion of posterior variances from Gaussian processes. 
This can be particularly advantageous in scenarios where computational resources are constrained. 
In such cases, calculating the variance may be deemed infeasible due to its inherent complexity, scaling as $\mathcal{O}(N^2)$, where $N$ is the size of the dataset. 
The aEIGP framework thus provides an effective solution by accommodating the constraints associated with computational resources without compromising the quality of predictive insights.






\section{Numerical Results}\label{subsec_Benchmarks}
Within this section, detailed information, including the experimental setup, code source, and related aspects of the numerical evaluation in \cref{sec_sim}, is provided. Additionally, further exploration of the subject matter is facilitated through the elucidation of additional simulation results.

\subsection{Simulation Configuration}
All simulations were conducted on an Apple MacBook Pro (16-inch, 2021) under Sonoma 14.12 version with the M1 Pro Chip and 16 GB of RAM. 
The code execution was carried out using Python under version 3.10.8.

\subsection{Additional Simulation Results}



\begin{figure*}[h!]
    \centering
    \subfloat{{\includegraphics[width=1\textwidth]{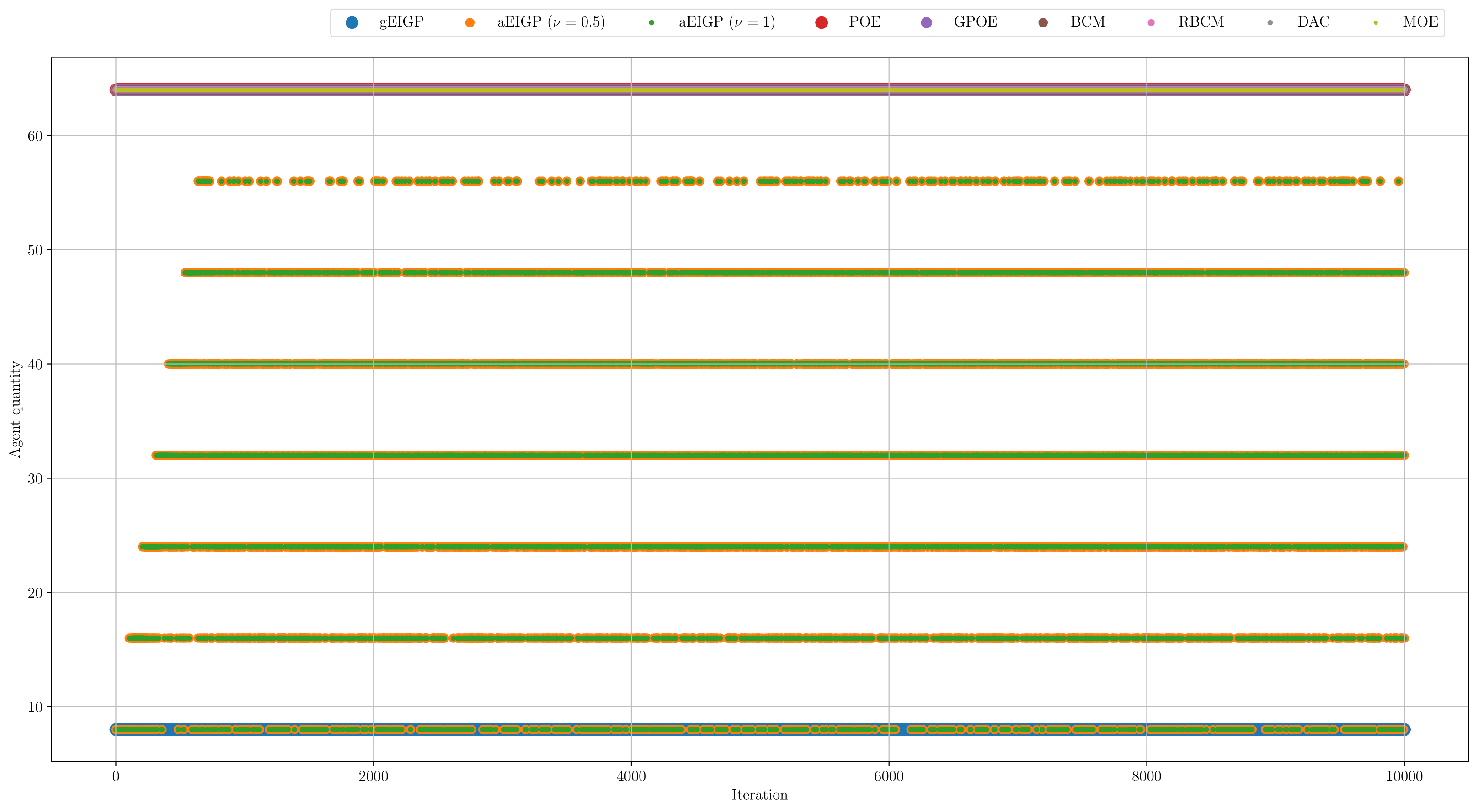} }}%
    \hspace{1cm} 
    \subfloat{{\includegraphics[width=1\textwidth]{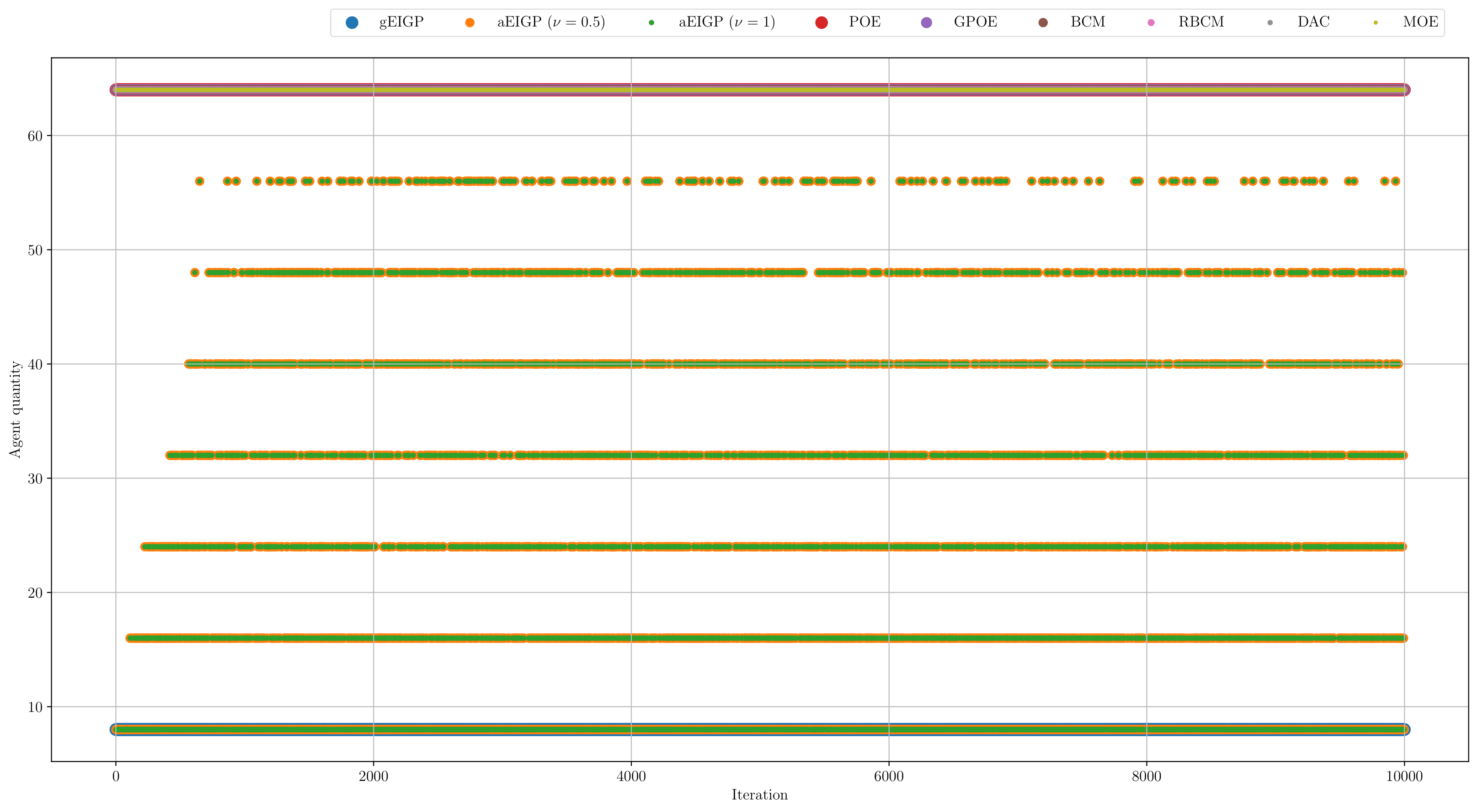} }}%
    \caption{The overall agent quantity within the MAS with 8 agents for each iteration in the KIN40K (top) and POL (bottom) data sets.}%
    \label{fig_kin40k_agentQuantity_8}%
\end{figure*}

\begin{figure*}[h!]
    \centering
    \subfloat{{\includegraphics[width=1\textwidth]{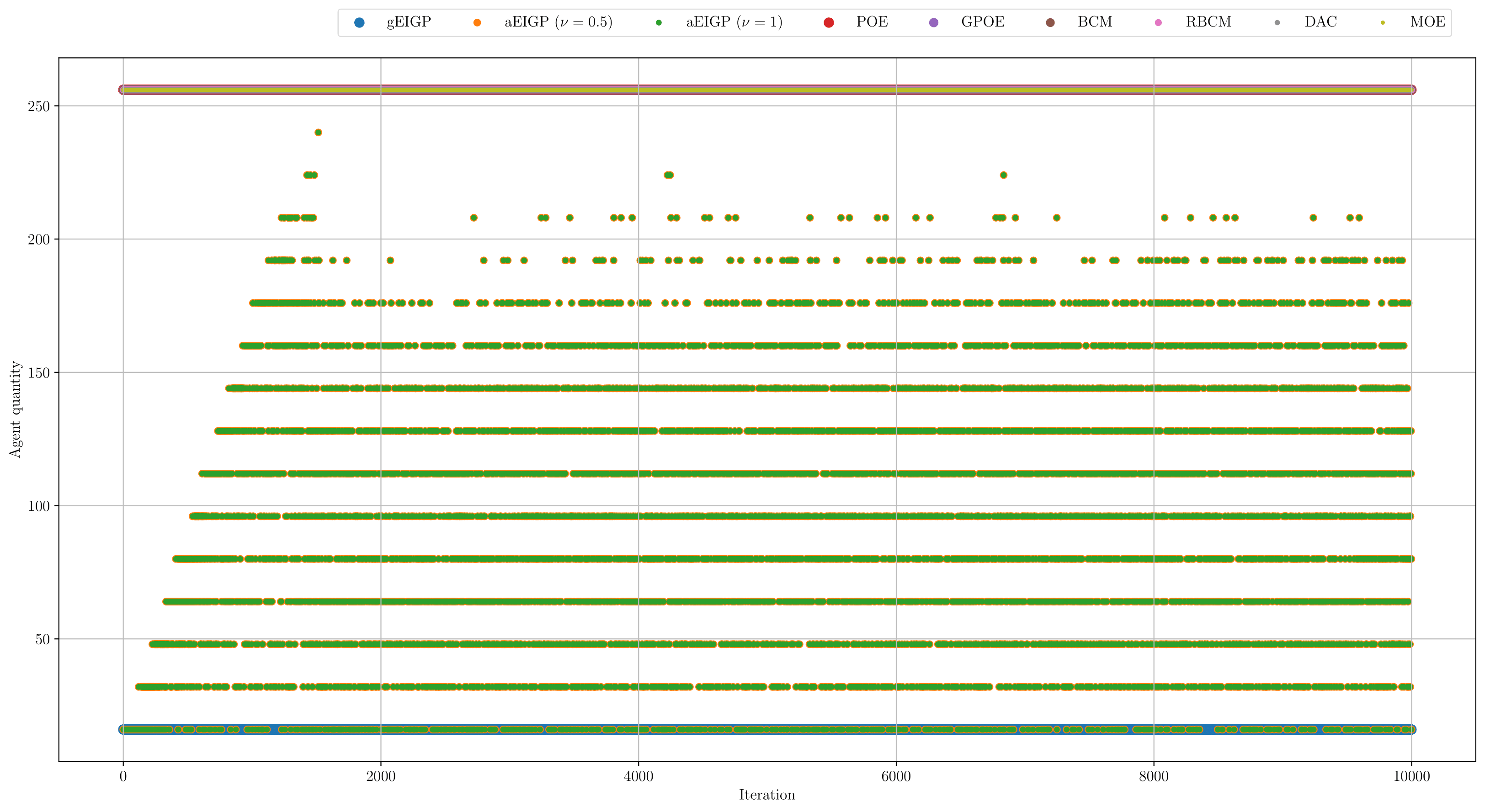} }}%
    \hspace{1cm} 
    \subfloat{{\includegraphics[width=1\textwidth]{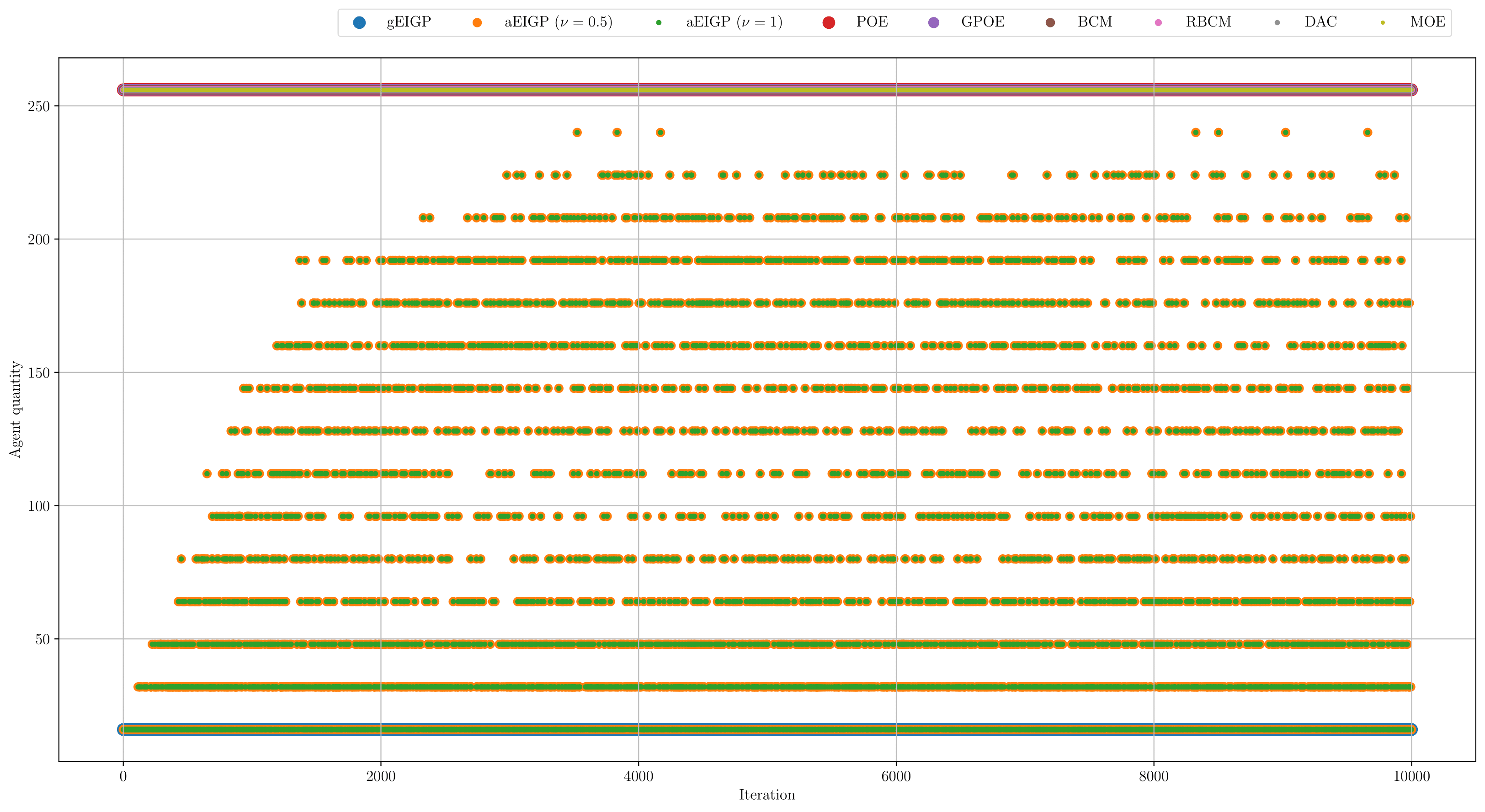} }}%
    \caption{The overall agent quantity within the MAS with 16 agents for each iteration in the KIN40K (top) and POL (bottom) data sets.}%
    \label{fig_pol_agentQuantity_16}%
\end{figure*}

From \cref{fig_kin40k_agentQuantity_8} to \ref{fig_pol_agentQuantity_16}, it is shown that the proposed EIGP methods demonstrate that not all agents need to be actively involved in making predictions, underscoring the importance of quality over quantity in cooperative learning.

\newpage
\bibliographystyle{IEEEtran}
\bibliography{ref}

%








\end{document}